# Optimizing Urban Critical Green Space Development Using Machine Learning

Mohammad Ganjirad[1], Mahmoud Reza Delavar[1, *], Hossein Bagheri[2], Mohammad Mehdi Azizi[3]

[1] Center of Excellence in Geomatic Engineering in Disaster Management and Land Administration in Smart City Lab, School of Surveying and Geospatial Engineering, College of Engineering, University of Tehran, Tehran, Iran; mohammad.ganjirad@ut.ac.ir; mdelavar@ut.ac.ir

[2] Faculty of Civil Engineering and Transportation, University of Isfahan, Isfahan, Iran; h.bagheri@cet.ui.ac.ir;

[3] School of Urban Planning, College of Fine Arts, University of Tehran, Tehran, Iran; mmazizi@ut.ac.ir;

[*] correspondence: mdelavar@ut.ac.ir ,

# Abstract

This paper presents a novel framework for prioritizing urban green space development in Tehran using diverse socio-economic, environmental, and sensitivity indices. The indices were derived from various sources including Google Earth Engine, air pollution measurements, municipal reports and the Weather Research & Forecasting (WRF) model. The WRF model was used to estimate the air temperature at a 1 km resolution due to insufficient meteorological stations, yielding RMSE and MAE values of 0.96°C and 0.92°C, respectively. After data preparation, several machine learning models were used for binary vegetation cover classification including XGBoost, LightGBM, Random Forest (RF) and Extra Trees. RF achieved the highest performance, exceeding 94% in Overall Accuracy, Recall, and F1-score. Then, the probability of areas lacking vegetation cover was assessed using socio-economic, environmental and sensitivity indices. This resulted in the RF generating an urban green space development prioritization map. Feature Importance Analysis revealed that the most significant indices were nightly land surface temperature (LST) and sensitive population. Finally, the framework performance was validated through microclimate simulation to assess the critical areas after and before the green space development by green roofs. The simulation demonstrated reducing air temperature by up to 0.67°C after utilizing the green roof technology in critical areas. As a result, this framework provides a valuable tool for urban planners to develop green spaces.

# 1 Introduction

According to World Bank reports, today more than 50% of the world's population, about 4.4 billion people, live in urban areas. By the end of 2050, this number is expected to double, and on average, 7 out of every 10 people will live in urban areas [1]. Rapid urbanization, although improving economic conditions and living standards, has led to numerous ecological, environmental and climatic problems [2, 3]. Urban population growth causes traffic problems, leading to longer commute times and exacerbated air pollution [4]. Moreover, replacing land covers with artificial materials such as asphalt and concrete, escalates the Urban Heat Island (UHI) phenomenon, where urban areas experience higher temperatures than other areas, leading to increased energy consumption and harm to the health of citizens [5]. In response to these challenges, the development of urban green spaces emerges as a crucial strategy. The development of green space is essential for maintaining ecological balance, reducing the impact of climate change and enhancing social and economic situation. However, this development is a complex issue influenced by various environmental, social and economic elements.

In urban environments, vegetation cover appears in various forms collectively referred to as urban green spaces. Urban green spaces are areas within cities predominantly covered with vegetation and accessible to the public. Examples include parks, gardens, green corridors, etc. [6]. Urban green space and vegetation cover are two related but distinct concepts in urban planning. This distinction makes prioritizing areas for green space development more challenging [7]. Each of these urban green space elements serves as a valuable tool for the well-being of residents, offering benefits such as improved air quality and enhanced mental and physical health, thereby contributing to an overall better quality of life for the inhabitants [8-10].

Numerous studies have examined the role of vegetation cover from different viewpoints, using diverse parameters and methods in an urban environment. For example, the study of urbanization in Mumbai, India, during the period 1997-2018, using Landsat data, showed a significant reduction in vegetation cover and double increase in the built-up area. The effect of this urbanization was seen through a 30.04°C gap between the land maximum and minimum surface temperature in 2018. On the other hand, regression analysis showed that UHI strength decreased by raising the Normalized Difference Vegetation Index (NDVI) [11]. Another study conducted in the Tokyo Olympic region employed the boosted regression tree model to see how the 2D and 3D parameters affect the Land Surface Temperature (LST). Gravity Park Index (GPI), Normalized Difference Built-up Index (NDBI), NDVI, and mean building height were the most important indices for this evaluation. The results demonstrated that the maximum effect of vegetation cover on mitigating LST was achieved at an NDVI value of 0.7, and a high density of vegetation cover was optional to achieve the minimum air temperature [12]. In another study conducted by Basu and Das on the city of Raiganj, the relationship between urban green spaces and LST was evaluated. This study utilized multi-scale geographically weighted regression (MGWR) and considered 18 parameters to analyze the intensity of the UHI effect over different years. The results indicated that a more significant presence of dense and continuous green spaces was more effective in mitigating UHI effects. However, urban expansion and the fragmentation of these green spaces reduced their effectiveness in decreasing LST levels [13]. Hou et al. ranked the environmental elements influencing the UHI effect [14]. In this study, four environmental indices including Vertical Aspect Ratio (VAR), Albedo, NDVI, and Night time Light (NTL), were extracted from 17 Chinese cities and used in the non-linear regression models such as the Random Forest (RF). In this study, NDVI and Albedo indices were the most important factors that could cause UHI.

Previous researches have shown that individuals residing in proximity to urban green spaces exhibit lower mortality rates, reduced stress levels, and lower incidences of violent behavior. Moreover, compared to those lacking access to such spaces, individuals living near urban green areas demonstrate heightened mood, attention, and physical activity levels [15, 16].

Green space development directly and indirectly influences the urban economy [17]. For instance, commercial sites near parks and green urban spaces often have higher prices and experience increased demand [18, 19]. Furthermore, the quality of urban parks is a crucial factor in attracting a multitude of tourists and enhancing the economic development of the area [20].

The allocation of urban green space should be conducted using a spatially equitable approach. For instance, regions below the poverty line or including minority groups often experience the greatest impact from adverse environmental conditions and heightened exposures compared to others [21, 22]. Therefore, ensuring the equitable distribution of vegetation cover can help communities facing the threat of environmental injustice in mitigating the various risks. In conclusion, the equitable distribution of green areas can serve as a tool for social justice by enhancing the quality of life, influencing land use patterns, and promoting health equity [23]. On the other hand, the planning for environmental justice has to be performed in a specific context that takes into account particular circumstances and local characteristics [24]. In other words, due to budgetary and human resource constraints, urban decision-makers must prioritize the development of different green regions to ensure optimum resource utilization.

Several studies have investigated the optimum places for developing vegetation to maximize its benefits using different techniques. One category of studies has utilized microclimate models to evaluate various scenarios for identifying optimum locations for urban green space development. For example, in a study conducted by MacLachlan et al., a new approach was presented to determine the optimum locations for tree planting using climate data and a 3D urban model to reduce UHI more effectively. In Perth, Australia, high-density and lower-density areas were considered, and different thermal conditions were investigated. The Solar and Long Wave Environmental Irradiance Geometry (SOLWEIG) model in QGIS software estimated the mean radiant temperature in 2008-2011, and various statistical analyses were implemented. The statistical analysis found that the regions with the highest density had a higher average annual temperature. Based on this, a specific site was selected for the simulation, and four scenarios were considered for planting 15 trees. The first scenario was based on real conditions, while the second scenario involved modeling according to the government development plan. The third scenario involved removing trees from the simulation, revealing the highest estimated temperatures. Finally, the last scenario involved placing trees in the hottest pixels. The simulation demonstrated that optimum placement of trees could reduce the temperature by 0.8°C across the site. This suggests that the strategic placement of trees is the most effective way to achieve the best vegetation cover transformation [25].

Another study evaluated the impact of urban green spaces on thermal comfort using the Envi-Met model in a residential block in Tabriz, Iran, employing various tree pattern scenarios. In the initial scenario, the current state of the study area was assessed, which included 153 trees. In subsequent scenarios, the number of trees was increased to 270 with an average spacing of 12 m, 405 with an average spacing of 10 m, and 540 with an average spacing of 9 m. Simulations based on these scenarios were conducted for both summer and winter days. The simulation results across various scenarios indicated that the third scenario (adding 405 trees) provided the best thermal comfort conditions in both summer and winter. Specifically, the average daily air temperature decreased by 0.29°C in summer and by 0.64°C in winter compared to the current state scenario [26].

Wang et al. evaluated the accuracy of microclimate simulations and the impact of tree cover ratio on urban air temperature. First, the performance of the Envi-Met model in simulating air temperature at a height of 1.5 m was validated using ground stations. The results demonstrated that the model accurately estimates air temperature with a $R^2$ of 93%. An increase in tree cover ratio across different scenarios proved a direct relationship between tree cover and cooling effects. For example, a 50% increase in tree cover led to a temperature decrease of 1.03°C [27].

Another category of studies has employed Multi-Criteria Decision-Making (MCDM) methods to identify optimum locations. For instance, a multi-objective spatial decision support framework was designed to prioritize tree planting locations in the Bronx, New York, to maximize multiple objectives. By using population census blocks and mathematical optimization techniques, 14 different scenarios were designed in order to determine the optimum locations for planting trees, taking into account the increase in tree cover, the reduction of $PM_{2.5}$, the reduction of the heat index, the reduction of the cost of planting trees, and the maximization of specific ecosystem services. The results showed that multi-objective decision-making could identify more suitable places to use the benefits of vegetation, and optimization frameworks could help urban planners develop vegetation from different perspectives [28].

A study conducted in Nanjing, China, utilized the Fuzzy Analytical Hierarchy Process (Fuzzy-AHP) to determine the best location for urban park development, considering 13 factors such as physical, environmental, accessibility, and human activity. The study found that only 5% of the available areas were highly suitable for urban site selection, and the most significant factors were carbon storage, NDVI, and the heat island effect [29].

Soil condition is another significant factor affecting vegetation cover development. A study was conducted in Ahvaz, Iran, to assess land suitability for green space development using the AHP method. The modeling process incorporated several parameters like soil type, chemical properties, soil characteristics, and topographic features. According to the results of AHP, 54.34% of the areas had severe restrictions on green space use [30].

In another study, the equity of green spaces in Tehran was examined using the Hybrid Factor and Analytical Network Process (F'ANP) method. Thus, in this study, only nine indices in 3 categories related to access, quality, and quantity of green space were produced, and the weight of the layers was calculated based on the percentage variance of each component. Although the F'ANP method does not rely on pairwise comparisons, unlike traditional approaches, it still faces limitations in handling many indices and analyzing complex relationships. In addition, it lacks the flexibility needed to incorporate new indices. The Gini coefficient results, used as a measure of inequality, showed that the imbalance in the quality (0.23) and quantity (0.38) components of green space was greater than that in accessibility (0.28), indicating an uneven distribution of these components across the study area [31].

Xu et al. evaluated the suitable locations for green roofs by integrating the Technique for Order of Preference by Similarity to Ideal Solution (TOPSIS) method with deep learning. In this approach, roofs were identified using a deep learning algorithm, and TOPSIS was employed to generate a prioritization map for the most suitable roofs. The results showed that distance from green spaces, population density, and air pollution levels had the most significant impact on the outcomes [32]. However, the performance of TOPSIS can vary, as the addition or removal of factors may change the final ranking [33].

Urban environments are entirely dynamic and complex and are constantly changing. Incorporating a wide range of elements into the modeling process for prioritizing urban areas in green space development and obtaining optimum spatial decisions is essential. Socio-economic indices play a vital role in this process, providing valuable insights into urban environments by modeling life activities flow, including information from economic activities, demographics, and social structures. Moreover, it is essential to note that socio-economic indices significantly impact the quality and quantity of green spaces. These elements reflect the preferences and needs of the public and influence the decisions made by the urban managers [34]. In other words, these indices assist planners in understanding the socio-economic conditions of different urban areas and effectively managing urban growth by anticipating future needs [35-37]. For instance, age structures and population density play a crucial role in green space development, which should be tailored to the demographic needs and specific age groups in different areas [38, 39]. Previous investigations also indicate a positive correlation between household economic activities and access to green spaces, with higher-income areas generally having larger and higher-quality green spaces [40, 41]. Besides the socio-economic indices, environmental parameters offer detailed insights into the state of the urban environment. This data is crucial for policy formulation, planning, and making informed decisions [42]. In more detail, environmental indices play a vital role in improving the prioritization of areas requiring green space development by offering critical insights into physical, natural, and ecological factors. Notable examples include indices such as carbon storage, NDVI, NDBI, LST, and others [43, 44]. For instance, urban parks are effective in mitigating the UHI effect, with those featuring denser vegetation, increasing the level of NDVI [29]. Furthermore, indices like LST and Albedo play a critical role in pinpointing areas most vulnerable to the UHI phenomenon [45].

Overall, integrating indices from different domains offers a holistic view that improves decision-making by considering environmental needs and social welfare while minimizing human sensitivity. Consequently, this approach makes urban green space development more sustainable, effective, and equitable.

A review of previous studies revealed that research in the field of green space development can be categorized into two major groups. The first group includes studies that rely on microclimate simulations and the evaluation of various scenarios to identify optimum locations for vegetation development [25, 26, 46-49]. Since microclimate simulations require access to highly detailed data, implementing microclimate models also demands substantial computational power [50, 51]. Therefore, the simulated areas in microclimate-based studies are typically very small, and limited to a local scale. In addition, in studies that utilize microclimate simulations, the identifying the optimum locations for vegetation development has been based solely on factors such as air temperature or environmental pollutants. More specifically, when using microclimate simulations, it is not feasible to incorporate additional parameters, such as socio-economic indices. Considering the limitations mentioned in previous studies, there is a need for a comprehensive approach to urban green space development that involves climatic and environmental parameters and incorporates socio-economic and sensitivity indices.

The second group of studies utilized MCDM methods and knowledge-based approaches to prioritize suitable locations for vegetation development [52-55]. In these methods, collecting expert opinions from various aspects and consolidating these opinions pose significant challenges [56, 57]. On the other hand, modeling the dynamic and complex environment of a megacity like Tehran, requires incorporating diverse criteria. This requirement often prevents experts from providing a thorough analysis that considers multiple factors. A literature review also confirmed that most studies have utilized a limited number of criteria, which can negatively impact the accuracy and generalization of the model results [58, 59]. Considering the limitations of previous investigations, a sophisticated framework should be designed for the optimum development of urban green spaces. This framework should seamlessly integrate a diverse set of socio-economic, environmental, and sensitivity indices while avoiding bias in the modeling process. Furthermore, it should be suitable for implementation in large and complex metropolitan areas like Tehran. On the other hand, managing the many indices in the urban environment is a complex challenge for human experts, and there are better solutions than relying on traditional methods. One solution is to use intelligent algorithms such as machine learning (ML) for the dynamic modeling of urban environments [60, 61].

Therefore, this study developed a framework based on 36 diverse indices using machine learning algorithms to generate a prioritization map for urban green space development at an appropriate scale. One of the primary methods for deriving various indices involves the implementation of remote sensing (RS) and Geospatial Information System (GIS). Using satellite imagery is a highly effective way of assessing suitable areas for green space development [62, 63]. Satellite imagery enables thorough monitoring of the Earth surface at regular intervals, facilitating temporal analysis. Different spectral bands make it possible to distinguish land use classes [64, 65]. Moreover, the accessibility of extensive satellite imagery archives enables the detection of trends in environmental change, making them powerful tools for the study [66-69]. On the other hand, GIS enables the organization, analysis, modeling and visualization of the spatial data. Therefore, integrating RS and GIS can discover more meaningful spatial relationships and patterns, which will help inform decision-making. In addition, developing cloud processing systems, such as Google Earth Engine (GEE), has given researchers the power of a computing system. Without local hardware, GEE can be conveniently used to perform complicated analyses on large datasets, allowing the study of long-term trends [70-73]. These systems can use vast amounts of data and explore their hidden relationships to solve problems [74].

Thus, the developed framework took advantage of integrating ML methods, GIS and RS technology. Integrating RS and GIS analyses using ML algorithms enhances spatial decision-making by improving efficiency, accuracy, and understanding of hidden spatial relationships. Finally, the effectiveness of the proposed framework in identifying critical areas was validated through microclimate simulations. To provide a practical solution, the impact of the green roof technology was also further evaluated as an additional experiment.

In summary, the developed framework involved the following contributions:

1. Employing and evaluating different ML classifiers for generating vegetation cover map.
2. Utilizing various socio-economic, environmental, and sensitivity indices to prioritize urban green space development by an intelligent model.
3. Assessing the importance of different features in urban green space development.
4. Evaluation of the effectiveness of the developed framework in prioritizing critical areas for urban green space development using microclimate simulation.
5. Assessing the impact of green roof technology on improving critical thermal conditions through a microclimate simulation.

The remaining sections of the paper are organized as follows: Section 2 introduces the study area and its characteristics. In Section 3, the primary data sources used in this research are described. Section 4 presents the main steps of the research methodology and proposes a framework for generating the urban green space development priority map. Section 5 provides the implementation results and discusses the effectiveness of the proposed framework. Finally, Section 6, Conclusion section summarizes the key findings and provides a concise of the research outcomes.

# 2 Study Area

The study area of this research is the city of Tehran. Figure 1 illustrates the geographical extent of Tehran, indicating the location of air quality monitoring stations and weather synoptic stations. Tehran sits on the southern slope of the Alborz Mountain range in northern Iran, spanning from 51° 4' to 51° 37' East Longitude and 35° 32' to 35° 52' North Latitude. Tehran is the most populated city in Iran, covering an area of around 730 square kilometers with a population of 8.6 million people as the night population of the city, which goes up to more than 12.5 million people during day where a huge number of populations residing outside Tehran move to the city for their jobs. Due to urbanization, the LST has increased significantly in Tehran. In the center of the city, the land use change has resulted in a temperature rise of 2°C to 3°C, while in the outskirts of the city, it has increased by 5°C to 7°C [75, 76]. Furthermore, Tehran is one of the most polluted cities in Iran in terms of air pollution. Both human activities, such as traffic and natural elements like being surrounded by the Alborz Mountain range, contribute to the exacerbation and sustained presence of pollutants [77, 78].

Finally, Tehran was selected as the study area of this research for various reasons, including its geographical extent, climatic diversity, population growth, intensive land use change and urbanization, and challenges of increasing temperature, air pollution issues, and complex urban patterns.

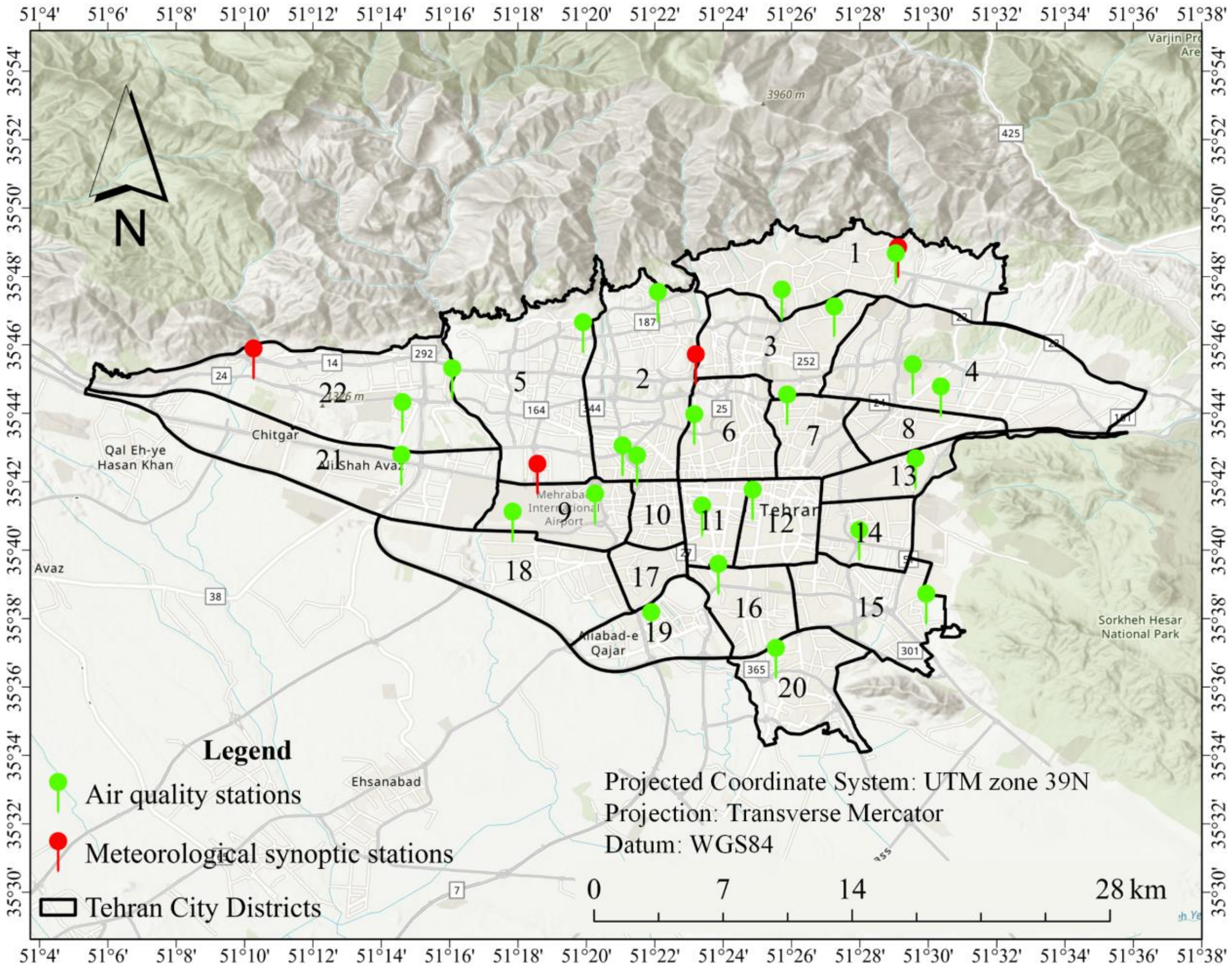

*Figure (1) Visualization of air pollution and meteorological synoptic stations in 22 Districts of Tehran.*

# 3 Materials

In this research, various data sources were used, including satellite imagery, meteorological and air pollution station data, and socio-economic information, to model the dynamic environment of Tehran. Table 1 lists all the indices, their abbreviations, source, temporal range, resolution, interpolation and, rasterization method. Additional details on obtaining the different indices are provided in Section 4.

*Table (1) The characteristics of different information layers used in this research.*

| Abbreviation | Layer Name | Source | Time | Resolution | Interpolation / |
|---|---|---|---|---|---|

| | | | | | Rasterization |
|---|---|---|---|---|---|
| ECP | Economic Participation | Tehran Municipality | 2022 | 100 m | Maximum area |
| FEP | Female Population | | | | |
| MP | Male Population | | | | |
| GA | Green Area | | | | |
| GBA | Green Belt Area | | | | |
| GC | Green Corridors | | | | |
| ER | Education Rate | | | | |
| PN | Normalized Parcel Numbers | | | | |
| SR | Sex Ratio | | | | |
| SP | Sensitive Population Rate | | | | |
| TER | Total Employment Rate | | | | |
| MER | Male Employment Rate | | | | |
| FER | Female Employment Rate | | | | |
| NDR | Net Dependency Rate | | | | |
| GDR | Gross Dependency Rate | | | | |
| LKC | The Number of Gardens Less than 1000 Square Meters | | | | |
| LKM | The Area of Gardens Less than 1000 Square Meters | | | | |
| L5KC | The Number of Gardens Between 1000 and 5000 Square Meters | | | | |
| L5KM | The Area of Gardens Between 1000 and 5000 Square Meters | | | | |
| L10KC | The Number of Gardens Between 5000 and 10000 Square Meters | | | | |
| L10KM | The Area of Gardens Between 5000 and 10000 Square Meters | | | | |
| L100KC | The Number of Gardens Between 10000 and 100000 Square Meters | | | | |
| L100KM | The Area of Gardens Between 10000 and 100000 Square Meters | | | | |
| M100KC | The Number of Gardens is More than 100000 Square Meters | | | | |
| M100KM | The Area of Gardens is More than 100000 Square Meters | | | | |
| $PM_{2.5}$ | 10-Year Mean Concentration of $PM_{2.5}$ | Air Quality Control Company | 21-03-2013 21-03-2023 | 100 m | Kriging |
| T2 | Air Temperature at 2 m above earth surface | WRF Model | 28-06-2022 | 1 km | Nearest Neighbor |
| WS | Wind Speed | | | | |
| DLST | Daily LST | MODIS (Terra) | 22-06-2022 22-09-2022 | 1 km | |
| NLST | Nightly LST | | | | |
| DIFFLST | Difference in Day and Night LST | | | | |
| NDVI | Normalized Difference Vegetation Index | Landsat 9 | | 30 m | |
| NDBI | Normalized Difference Built-Up Index | | | | |
| EVI | Enhanced Vegetation Index | | | | |
| SAVI | Soil Adjusted Vegetation Index | | | | |
| NDMI | Normalized Difference Moisture Index | | | | |

## 3.1 Satellite Dataset

Landsat series are among the most well-known and widely used sensors in various research [79, 80]. Landsat 9, the latest addition to the Landsat constellation, was launched into orbit on September 27, 2021. It is outfitted with two primary instruments: the Operational Land Imager 2 (OLI2) and the Thermal Infrared Sensor 2 (TIRS2). OLI2 provides imagery across up to 9 spectral bands in the visible and near-infrared spectrum. This study obtained surface reflectance images using the Landsat 9 sensor available in the GEE. These images were utilized to compute various spectral indices, with a maximum of 10% cloud cover, during the period from June 22, 2022, to September 22, 2022, resulting in the processing of 9 images. Landsat 9 images are captured at intervals of 16 days, which may not be optimum for monitoring dynamic changes like LST. To address this limitation, MODIS sensors aboard the Terra satellite have been employed to derive LST. MODIS offers 36 spectral bands with resolutions ranging from 250 to 1000 m which operates with an average temporal resolution of 1 day. In addition, the time pass of the MODIS on the Terra satellite is approximately 10:30 AM and 10:30 PM local time [81]. Given these considerations and the fact that the quality and accuracy of MODIS sensor products have been validated in numerous studies, its products were selected for use in this research [82-84]. Thus, among June 22, 2022, and September 22, 2022, 92 daily and nightly LST products were processed using GEE. It is important to note that analyzing thermal data during both day and night can improve the understanding of thermal behaviors and identify temperature anomalies.

## 3.2 Socio-economic Dataset

It is essential to include socio-economic indices in addition to environmental parameters to prioritize urban areas in green space development. Socio-economic factors play an important role in the background and influence environmental changes. Socio-economic indices contain a broad range of features that reflect various aspects of society, economy and can assist understanding the behavior of citizens more precisely [85]. These parameters provide insights into government decisions and people's demographic, economic, social, and health conditions, which can be very valuable for future decision-making in urban planning, vegetation management, and environmental justice [86-89]. Therefore, in addition to environmental parameters, socio-economic indices were considered for the study area. These indices include sensitive population, population (male, female), economic participation rate, sex ratio, employment rate (male, female, total), net and gross dependency ratio, education rate, and parcel numbers (Table 1).

## 3.3 Urban Green Space Dataset

Urban green spaces are crucial for the environment, society, and economy. They help reduce air pollution and UHI, increase physical and mental activity, and improve the overall well-being of residents [90-92]. In addition to socio-economic indices, parameters related to vegetation cover and urban green spaces were considered across various classifications. These parameters identify areas with low vegetation cover, enabling prioritization for green space development.

## 3.4 Meteorological and Air Pollution Dataset

Air pollution is a significant concern for public health. In Tehran, $PM_{2.5}$ is the primary pollutant based on official reports [93-95]. The daily $PM_{2.5}$ concentration data collected from 24 Air Quality Control Company (AQCC) stations in Tehran have been used to calculate the 10-year mean concentration. However, it is essential to note that the meteorological conditions of a particular area strongly influence air pollution [96]. In other words, the meteorological conditions of a region can significantly affect the dispersion, concentration, and transformation of various pollutants. For this reason, the WRF physical model was employed to simulate the climate situation of the study area and its effect on modeling green space development.

# 4 Methodology

This section describes the essential steps of the proposed framework to identify priority areas for green space development. This framework allows urban managers and planners to identify and prioritize urban green space development areas by integrating various socio-economic, environmental, and sensitivity indices. Figure 2 illustrates the primary steps of this study. This framework utilizes the inherent characteristics of the data and employs advanced machine learning algorithms to reveal hidden relationships among the variables, thus reducing bias in decision-making. Initially, areas with and without vegetation cover are accurately classified, and ultimately, a green space development prioritization map is generated using a diverse set of indices. In addition, microclimate simulations are conducted to validate the results and evaluate the impact of developing green spaces in high-priority areas, specifically through green roofs. The proposed framework consists of three main part including data collection and preprocessing, classification to determine vegetation cover status, spatial analysis, and decision-making for prioritizing urban green space development. In the first step, the required data were collected. The data include socio-economic indices derived from various GIS layers, meteorological station observations of air temperature, 10-year station observations of $PM_{2.5}$ concentration, different spectral indices, inputs for the WRF physical model, and finally, collected labels for classification to classify vegetated and non-vegetated areas. Then, preprocessing was performed on each dataset, and information layers were used to identify areas with higher priority for green space development. At the beginning of this stage, a correlation analysis was conducted on the information layers, and effective features were selected. Then, the classification process was performed on the final dataset using ML algorithms such as eXtreme Gradient Boosting (XGBoost), Extra Trees (ET), Random Forest (RF), and Light Gradient-Boosting Machine (LightGBM). In this framework for prioritizing green space development, the initial step involves identifying vegetated and non-vegetated areas, and then priority allocation must be assigned to non-vegetated areas. The best ML model was selected based on the evaluation results, and then a binary vegetation cover map was generated. While the binary map distinguishes between vegetated and non-vegetated areas, it lacks additional information. In other words, not all non-vegetated classes hold equal importance for development; thus, they necessitate ranking and prioritization based on other indices. In simpler terms, these areas must be prioritized, and areas requiring green space development (high priority) and areas with low priority must be determined. The prioritization process began with re-feature selection, where indices only useful for classifying vegetation cover, such as NDVI, were removed from the dataset. Afterward, a probability map was generated using the remaining features to identify areas that require green space development. This probability map predicts which pixels are in a critical state due to lack of vegetation. The probability and binary maps were integrated to generate a development map of green space for equitable growth. Finally, the importance of features and their role in the probability map was evaluated. Each of these steps is explained in more detail in the following sections.

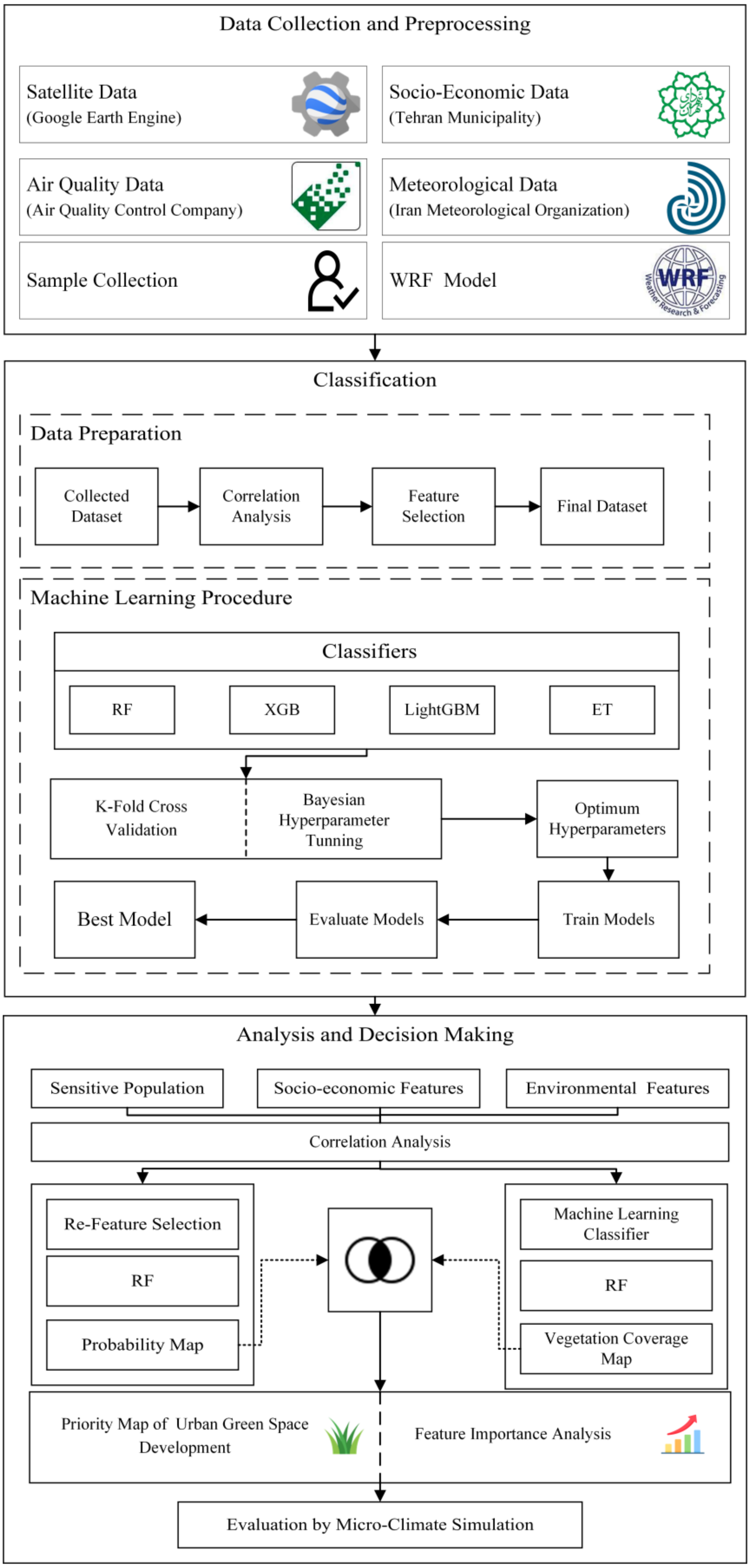

*Figure (2) The designed framework to identify priority areas for green space development, considering various parameters, using ML techniques.*

## 4.1 Data Collection and Preprocessing

In this study, various indices and information layers were generated. Significant inconsistencies were observed among the initial datasets sourced from multiple providers. Addressing these inconsistencies is essential to mitigate sources of uncertainty, highlighting the critical importance of preprocessing in creating a unified and consistent dataset. To ensure the robustness of the preprocessing stage, the accuracy and quality of each input index were thoroughly evaluated using available resources. This step was undertaken to minimize potential modeling errors and prevent their propagation throughout the analysis.

Notably, according to reports from the Iran Meteorological Organization, June 28, 2022, with a temperature of 42°C, was the fourth hottest day in the past decade. Tehran experienced one of the hottest summers on record. Therefore, satellite-based indices were calculated during this hot summer period in addition to the socio-economic indicators for that year.

### 4.1.1 Spectral Indices

One of the primary goals of this research is to classify vegetated from non-vegetated areas. Various spectral indices play a crucial role in extracting valuable information [97]. Several studies have shown the performance of different spectral indices as a powerful tool in distinguishing classes and increasing classification accuracy [98-101]. For one of the hottest summers in Tehran history, from June 22, 2022, to September 22, 2022, the mean indices, including Soil Adjusted Vegetation Index (SAVI), Normalized Difference Built-up Index (NDBI), Normalized Difference Moisture Index (NDMI), Normalized Difference Vegetation Index (NDVI), and Enhanced Vegetation Index (EVI), were calculated using the GEE platform. Landsat 9 surface reflectance images during the specified period were collected to compute indices, ensuring a cloud cover of less than 10%.

### 4.1.2 LST Indices

LST is one of the most crucial parameters for prioritizing urban green space development. LST is entirely related to the urban morphology [102]. With the increase in population, rapid changes in land cover take place, which in turn leads to increases in impervious surfaces and subsequently raises the LST and the UHI effect [103]. Thus, a LST-based analysis will obtain temperature patterns in different urban areas and identify the critical regions affected by the UHI effect. Few sensors produce nightly LST products compared to daily LST. However, LST at night is crucial for human thermal comfort and indirectly affects nightly atmospheric pollution [104]. Also, according to a previous study, in semi-arid regions, nightly LST maps show UHI patterns more clearly than daily LST [105].

On the other hand, the differences between nightly and daily LST are essential information that indicates the intensity variations, which can be attributed to the UHI effect. The mean daily and nightly LST and their absolute differences were extracted from the MODIS sensor on the Terra satellite using the GEE system, covering the period from June 22, 2022, to September 22, 2022. During the preprocessing stage, the nearest neighbor interpolation method was used to enhance the spatial resolution of these datasets from 1 km to 100 m, resulting in a consistent and integrated final dataset.

### 4.1.3 Socio-economic and Urban Green Space Layers

Tehran Municipality collected the raw data about socio-economic and urban green space indices in Excel file format. The dataset was preprocessed before entering the modeling process. This involved writing the information for each of the 22 Tehran Districts (Figure 1) as an attribute in a Shapefile. The data was then rasterized using the nearest neighbor method, where each raster cell represented the value of that index in the corresponding district.

### 4.1.4 Sample Collection

As mentioned before, it is necessary to classify vegetated and non-vegetated areas to prioritize green space development. This process was performed under supervised classification. The optimum number of samples with proper spatial distribution is essential to achieve a high-quality map in the classification process. In this study, 4832 samples were extracted from high-resolution satellite imagery using Google

Earth Pro software during the summer of 2022 by two experts. The experts collected samples by considering environmental conditions, socio-economic development levels, and sensitivity statuses across different regions. Specifically, the experts collected samples labeled as ‘1’ from areas identified as critical based on multiple circumstances where the necessity for vegetation development was evident. In contrast, samples labeled as ‘0’ were gathered from regions deemed non-critical across various parameters, indicating that vegetation development was not a priority in these areas. This classification underscores the urgency of green space development in the identified critical regions. It is essential to have an equally distributed data set to prevent the model bias toward the majority class, reducing the chances of overfitting, selecting the most critical features, and improving the model accuracy and interpretability [106-108]. This study conducted balanced data extraction, resulting in 2788 samples for class 0 and 2044 samples for class 1.

#### 4.1.5 Air Quality Data Processing

Air pollution is a significant issue in Tehran, which constantly threatens the physical and mental health of citizens [77, 109]. As a result, it is vital to include indices related to air pollution in prioritizing the development of green space. In other words, this parameter enhances the modeling process by reflecting the impact of the environment quality on the health of the citizens. This research calculated the mean concentration of $PM_{2.5}$ for ten years using raw observational station data. Evaluating ten years of data compared to shorter periods provides a comprehensive view of air quality trends and provides more reliable and accurate information. This means that pollutant concentration fluctuations due to temporary reasons such as fire, industrial accidents, or epidemic diseases such as COVID-19 become less effective, and more precise information on the main trend is obtained. Daily data from 24 stations operated by the Tehran Municipality Air Pollution Quality Control Company, spanning from March 21, 2013, to March 21, 2023, was utilized for the analysis. To avoid error propagation and maintain the quality of the final map, days with fewer than 10 observation stations were excluded from the dataset through a filtering process. In addition, for interpolation, since Simple Kriging has shown better results in previous studies for Tehran, this method was employed along with a spherical semi-variogram [110]. Furthermore, to ensure more accurate interpolation, optimum hyperparameters were determined using cross-validation to minimize interpolation error. Finally, all the generated layers were averaged to produce the map representing the 10-year average $PM_{2.5}$ concentration.

### 4.1.6 WRF Simulation for Temperature and Wind Speed Estimation

Meteorological parameters such as temperature and wind speed are major elements for detecting vegetation coverage. Plants have different temperature requirements, and extreme temperatures and heat stress can cause vegetation degradation. Thus, the increased usage of more resilient vegetation cover in high-average-temperature regions is suggested [111, 112]. Moreover, high temperatures can pose significant health risks to citizens. For example, in China, elevated temperatures have been linked to increased mortality rates due to respiratory diseases, strokes, and circulatory disorders [113]. Wind speed also plays a significant role in plant transpiration, with high wind speeds often leading to increased water stress [114]. In addition, wind speed affects the temperature, humidity, and air pollution [115-117]. Utilizing meteorological parameters in conjunction with other indices provides a more comprehensive approach to prioritizing areas for green space development. However, unlike air pollution stations, the number and distribution of meteorological synoptic stations are inappropriate over the study area. This means that the distance between stations is vast, and using the interpolation method to produce layers related to meteorological parameters is ineffective and does not accurately reflect the current situation. Consequently, the WRF model was employed to derive the necessary meteorological layer.

WRF is a numerical weather prediction system used for operational forecasting and atmospheric science research [118-120]. The simulations of this system are based on physical parameters at different scales. In addition to estimating meteorological parameters, the WRF model has been developed for various applications using different extensions. These include WRF-Chem for atmospheric chemistry, WRF-Hydro for hydrology, and WRF-Fire for wildfire simulation. Many researchers have validated its results across these domains [121-125]. In the WRF modeling system, two sets of input data, namely WRF Terrestrial Data and WRF Gridded Data, are initially introduced into the WRF Preprocessing System (WPS). The WRF Terrestrial Data includes land use data, surface albedo, topography, soil type, etc., which are considered static layers and undergo minimum changes over time. On the other hand, WRF

Gridded Data includes meteorological data that is dynamically incorporated into the modeling process. These meteorological data are obtained from the output of global models such as the Global Forecast System (GFS), European Centre for Medium-Range Weather Forecasts (ECMWF), etc., for each simulation in the model. First, the WPS section determines the projection system, domains of simulation, and their resolutions. Then, the aforementioned input data are merged into the defined domain using the interpolation method. Then, the Real program computes the initial and boundary conditions from the integrated dataset to solve various physical schemes. Finally, the Advanced Research WRF (ARW) core is employed, leveraging various numerical methods to solve the set of physical equations throughout the simulation time frame. These methods discretize the physical equations in space and time, providing estimates of meteorological parameters in short steps [126, 127].

In order to implement the WRF model in the study area, appropriate configurations such as the projection system, simulation period, and land use scheme must be determined. This research employed the Lambert conformal conic projection system centered on Tehran. As depicted in Figure 3, three nested domains were defined with resolutions of 1 km, 3 km, and 9 km. All static data were utilized at the highest resolution. In addition, the Modis_15s_Land use map was integrated into the modeling. Meteorological data from the Global Data Assimilation System/Final Analysis (GDAS/FNL), provided by the National Centers for Environmental Prediction (NCEP), was employed with a resolution of 0.25 degrees. The primary objective of implementing the WRF model was to estimate meteorological parameters on the hottest day of the year (June 28, 2022). To minimize uncertainty, the meteorological data from June 26 to June 29, 2022, were considered after a two-day spin-up time. The task of selecting the most appropriate set of schemes, such as soil, radiation, and others, is extremely challenging. Thus, this requires a number of model executions and appropriate knowledge of atmospheric physics. Fortunately, with Version 3.9 of WRF, a physics suite option is now available with validated schemes showing reasonable and good results. According to the previous research, the CONUS physics suite was used in this case study [128-130]. It is worth noting that the model execution was performed in parallel using an Intel® Core™ i7-12700K CPU with 20 logical processors.

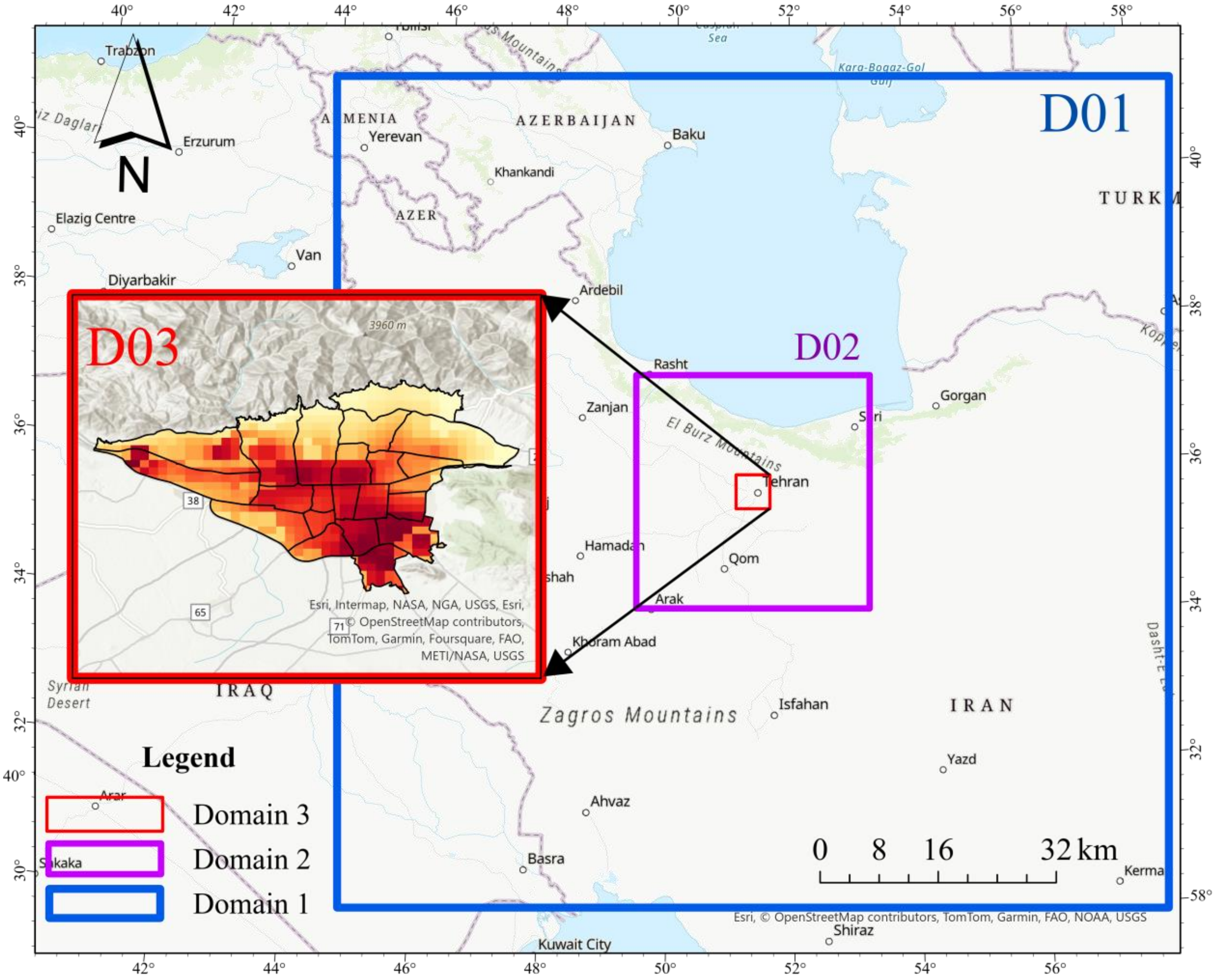

*Figure (3) Defined nested domains of the WRF model on the study area.*

## 4.2 Smart Decision Making based on Machine Learning

The primary goal of the designed framework is the intelligent prioritization of urban areas that require green space development based on various environmental, socio-economic, and sensitivity indices. The "Analysis and Decision-Making" part of Figure 2 illustrates the main steps for generating the priority map of green space development. To achieve the above goal, the first step is to identify vegetated and non-vegetated areas and then intelligently prioritize them, mainly focusing on areas lacking vegetation cover. The initial step of this framework involves binary classification of vegetated and non-vegetated areas without any additional information. Nevertheless, green space development is affected by various factors, such as environmental conditions, socio-economic behaviors, distribution of green spaces, and land management practices. To make informed decisions concerning critical areas, it is crucial to prioritize them based on different parameters and circumstances. By integrating various indices, the generated priority map will comprehensively involve factors that influence green space development. This map will help identify optimum areas for green space development by allocating limited resources like budget, workforce, and time. In general, while a binary vegetation cover map serves as a fundamental tool for identifying vegetated areas, a prioritization map that integrates all relevant indices offers enhanced accuracy and provides a comprehensive perspective, enabling more informed and sustainable decision-making.

As shown in Figure 2, the raw dataset is first passed through a classification process employing different ML algorithms, which results in a precise binary vegetation cover map. Subsequently, to prioritize and identify critical non-vegetated areas, re-feature selection is performed, removing features valuable only for binary classification, such as NDVI, and creating a new dataset. This approach enhances the role and importance of other environmental, socio-economic, and sensitivity indices within the ML model. By removing features relevant only to binary classification, such as NDVI, the model must rely on other indices to classify and uncover hidden patterns. The model is subsequently trained on this new dataset, generating a probability map determining the likelihood that a pixel is non-vegetated. In other words, by utilizing the re-feature selection approach, the model assesses the criticality of each pixel about the absence of vegetation. Finally, the probability map and binary vegetation cover map are integrated to obtain a priority map of green space development in a continuous range of 0 to 1.

### 4.2.1 Correlation Analysis

Correlation analysis is essential to discover the effective features in the data preprocessing stage. Features that exhibit high similarity do not contribute additional useful information to the model; instead, they cause redundancy in data. This redundancy may skew the model performance towards specific features which results in a bias towards particular classes during the classification process. Moreover, a large number of features complicates the interpretation of model performance. Employing correlation analysis generally enhances accuracy and efficiency [131, 132]. In this study, the correlation matrix was utilized to evaluate similarities.

### 4.2.2 Binary Classification for Recognizing Vegetation Coverage

After removing redundant features, the prepared dataset was used to classify vegetated and non-vegetated areas. In this step, different ML algorithms were evaluated, and the best one was selected. The research was carried out using four models, namely RF, LightGBM, XGBoost and ET. RF relies on an ensemble of decision trees for classification, with each tree producing individual outputs combined to yield improved results. RF is particularly effective in handling complex datasets and is known for preventing overfitting [133]. XGBoost is a gradient-boosting algorithm known for its high computational speed and efficiency. In this algorithm, decision trees are created sequentially, and each subsequent is trained to correct the previous model errors [134, 135]. The advantage of this method lies in the sequential training approach, which mitigates the risk of all models making the same mistakes. Like RF, ET creates a forest of random trees, while split points are randomly selected. This means trees are generated more randomly than in RF, and ET has shorter training times [136]. In addition, LightGBM is a gradient-boosting framework developed by Microsoft that is highly efficient and scalable when working with large datasets. In the LightGBM algorithm, trees grow vertically by adding leaf nodes, while in the other algorithms, tree growth is in-depth or level relative to the leaf. LightGBM is also very light and fast [137, 138].

### 4.2.3 Hyperparameter Tuning

The optimization of hyperparameters is a key element in the performance of ML algorithms [139]. The Bayesian approach was used to tune the hyperparameters in this research. In the random and grid search methods, previous results are completely ignored, and the model starts to search for parameters in the whole range again. Nevertheless, the Bayesian method evaluates past results and constructs a probabilistic (surrogate) model to approximate the objective function. This function aims to determine the next hyperparameters by creating a balance between exploring new hyperparameter spaces and exploiting previously evaluated areas. This process continues until the terminating criterion is met. Table 2 displays the optimum hyperparameter identified for different models after 20 iterations.

*Table (2) The set of hyperparameters obtained for each ML model after undergoing the Bayesian hyperparameter tuning process.*

| Model | Best Hyperparameters |
|---|---|
| RF | • Number of trees = 400<br>• Max depth = 5<br>• Max features = 0.6283<br>• Max samples = 0.9471 |
| XGBoost | • Number of trees = 2000<br>• Subsample = 0.8734<br>• Max depth = 17<br>• Min child weight = 5<br>• Reg lambda = 0.3346<br>• Eta = 0.0122<br>• Gamma = 7.4644<br>• Col sample by tree = 0.7473 |
| LightGBM | • Number of trees = 800<br>• Number of leaves = 2<br>• Boosting Type = dart<br>• Learning rate = 0.0805<br>• Max depth = 17<br>• Reg lambda = 0.3659<br>• Subsample=0.8789<br>• Col sample by tree= 0.7394 |
| ET | • Number of trees = 400<br>• Max depth = 5<br>• Max features = 0.6283<br>• Max samples = 0.9471 |

### 4.2.4 Accuracy Assessments of Classifiers

The performance of various ML models in classifying vegetated and non-vegetated areas was evaluated using different metrices such as Overall Accuracy (OA), Precision, Recall, and F1-score. These metrices provide a comprehensive understanding of the classification performance by considering various aspects.

### 4.2.5 Priority Mapping for Non-Vegetated Areas

The output of the classification is a map that identifies non-vegetated areas. However, not all of these areas are suitable for green space development. The primary question arises: which of these pixels are more critical and have a higher priority for green space development? In addition, how can the priority of these critical areas be determined using the available dataset?

Since the collected samples are derived from high-resolution images mainly focusing on vegetation, the binary vegetation classification model is highly sensitive to the NDVI index. In order to prioritize the regions, the role and importance of other parameters, such as socio-economic, environmental, and sensitivity indices, should be highlighted in the dataset. Indeed, after generating the binary map, the NDVI index, which is primarily valuable for vegetation detection, no longer offers additional information for prioritizing critical areas. By eliminating NDVI, the ML model must perform based on the other indices, enhancing its ability to determine intricate relationships and patterns. By adopting this strategy, the model could estimate each pixel criticality regarding the absence of vegetation cover using other environmental, socio-economic, and sensitivity indices.

To leverage the valuable information from the NDVI and other indices, first, the binary classification was generated the vegetation cover map with high accuracy by the dataset, including the NDVI. Then, the process of re-feature selection was performed. The NDVI index was removed from the data set due to the high sensitivity of the model, which made the importance of other indices more prominent. The RF was used to predict the probability of each pixel belonging to the non-vegetation class. Finally, a priority map of green space development was generated by integrating these two output maps. In this combination, zero classes (representing vegetation covers) remained constant, while only the values of non-vegetated areas (class one) were updated using the second map. This designed framework ensures using the valuable information of NDVI while enabling the model to uncover more diverse patterns within the new dataset based on the features.

## 4.3 Validating Urban Green Space Prioritization by Microclimate Simulation

After preparing the green space prioritization map, the first step is to validate the performance of the developed framework. Thus, the critical conditions of various areas and potential improvement strategies should be examined in more detail. It should be noted that Tehran has a highly complex and dynamic environment, with each area having its own specific limitations. The existing conditions of each critical zone vary significantly from different perspectives, such as urban architecture, availability of open spaces, building materials, etc. Given these conditions, the use of microclimate simulation presents an effective practical solution for understanding the current state and serves as a powerful tool for evaluating various scenarios. In this study, after identifying the critical areas, one of these regions was selected for microclimate simulation using the Envi-Met software. Figure 4 illustrates the main processes involved in this simulation. As depicted in Figure 4, the necessary data were collected through on-site visits to the critical area. The collected data include the average elevation above mean sea level, the locations and types of trees, and the heights and positions of buildings. In addition, detailed information about the building materials, construction types, and soil composition of the selected area was gathered to ensure comprehensive data for analysis. After gathering the required dataset, the selected area was modeled using the Envi-Met software. Using the software database, the characteristics of trees, buildings, and surfaces were assigned for accurate simulation. This study considered two scenarios for the critical area, as illustrated in Figure 4. The first scenario represents the critical area state, providing a basis to validate the proposed framework effectiveness in accurately identifying critical zones. Furthermore, to evaluate potential solutions for improving critical conditions, the second scenario examines the impact of green roof technology. Specifically, given factors such as the lack of suitable open spaces and the high land cost in the selected area, green roofs were considered as an optimum solution for mitigating the issues in critical zones. After configuring the various scenarios, simulations were conducted on the hottest day of the year (June 28, 2022), starting at 6:00, and running for 18 hours. The Envi-Met model provides BIO-Met and Leonardo modules to visualize and analyze results. The results were then extracted and thoroughly analyzed to evaluate the microclimate condition before and after the green space development.

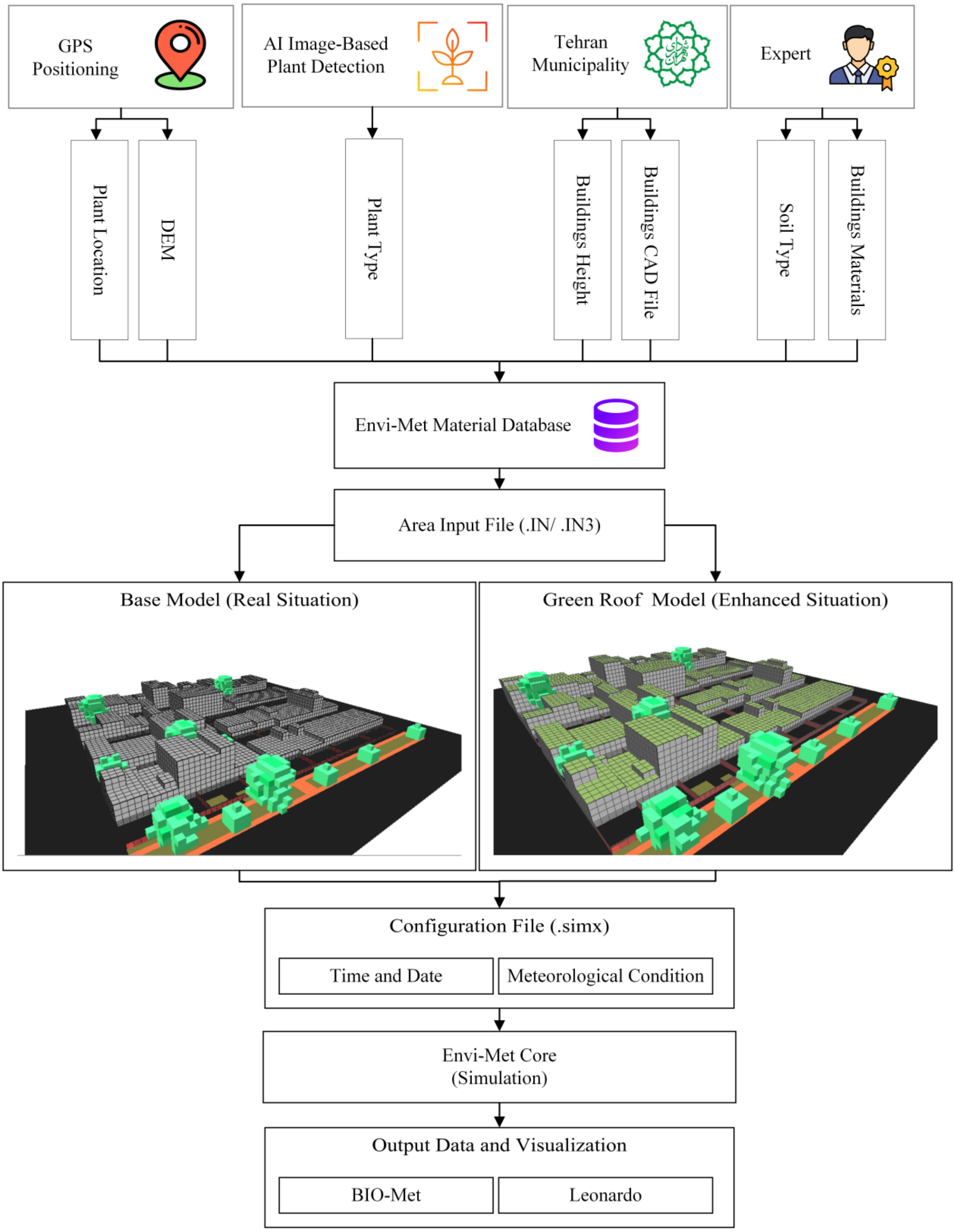

*Figure (4) The main steps for microclimate simulation in one of the identified critical areas.*

# 5 Results and Discussion

## 5.1 The Results of Estimating Socio-economic Parameters

After appropriate preprocessing, various socio-economic indices of Tehran were produced as maps (Figure 5). In Figure 5a, the sensitive population map reveals the proportion of the sensitive population, which includes children under 15 years old and the elderly over 65 years old. Elderly individuals living independently face various challenges, including inadequate housing, loss of residents' comforts, increased fear of crime, etc. Consequently, the quality of life for the elderly must be maintained, which requires urban planners to have sufficient information about their living environments [140]. In other words, analyzing this information will significantly contribute to create age-friendly urban environments and providing specific services to meet their unique needs [141]. Parks are among the most effective components in age-friendly cities. According to research that was conducted in five cities, including San Francisco, Houston, Atlanta, Chicago, and Boston, among people over 55 years old, it was found that visiting the park is important for more than half of the older adults [142]. It is important to consider the elements that facilitate access and use of parks for the population's physical and mental health benefits [143].

Children are also recognized as another part of the sensitive population, and their quality of life is crucial for the growth and future of society. Like the elderly, children are more sensitive to environmental hazards such as air pollution, heat waves, traffic congestion, etc., due to their weaker immune systems. A study conducted between 1994 and 2000 found increased hospital admissions for children and the elderly in London during heat waves [144]. In addition, providing sufficient opportunities for physical activities and entertainment is essential for the healthy growth of children. Access to parks, green spaces, and playgrounds reduces sedentary behaviors. A study conducted in the US found that children aged 0-17 living near parks had better physical and mental health. Green spaces were identified as a low-cost strategy for improving children's health conditions [145]. Consequently, interventions to improve environmental quality and increase green infrastructure in areas with highly sensitive population rates are necessary to mitigate the adverse effects of urbanization and pollution. For the aforementioned reasons, the sensitive population rate of the study area was extracted and illustrated in

Figure 5a. This map shows Districts 10, 14, and 17 have the highest sensitive populations. Considering the normalized male and female population maps, it can be observed that these districts also have the highest overall populations, and this high population density leads to increased pressure on resources and infrastructures. Furthermore, central areas experience higher pollution levels than other areas due to traffic and denser buildings, resulting in higher environmental stress.

The next parameter is the economic participation rate, which is calculated based on the concept of the active population. All individuals aged 10 years and above who were engaged in the production of goods and services (employed) or had the capacity to participate (unemployed) during the calendar week before the census (reference week) are considered economically active. The economic participation rate is calculated as the ratio of the active population to the total population aged 10 years and above. A higher economic participation rate indicates economic stability, increased activity, and investment potential, allowing more funds to be invested in increasing green space and planting trees. In the corresponding map, Districts 10, 15, and 19 exhibit the highest economic participation rates. These districts generally have a more stable economy, making them more flexible to economic fluctuations. This flexibility ensures the stability of green space development projects.

Another important parameter is the education rate or the percentage of adults who can read and write. Regions with higher education rates, such as Districts 3, 7, and 13, are more likely to support and participate in green space development projects. In other words, they are more aware of environmental issues and challenges and are more interested in implementing sustainable development methods.

The sex ratio is another socio-economic index. It is identified as the number of males per female. The sex ratio can offer valuable insights into the social dynamics of a community. If the sex ratio is imbalanced, community members may have different priorities and perspectives, influencing the community's involvement in green space development projects.

Another important parameter is the employment rate. According to official documents, the employment rate includes all individuals aged 10 and above who have worked for at least one hour during the reference week or have temporarily left their jobs. These individuals are considered employed. Districts with higher employment rates, such as 7, 12, 15, and 19, tend to have more economic activity. This issue can affect land use patterns and lead to their change. Air and noise pollution tends to be high in areas with high employment rates. Such regions also face increased demand for water and energy resources, which may affect green space development plans. Also, in these regions, there is a limitation in the availability of sufficient space for traditional green space development. Because these areas generally include buildings and transportation networks. On the other hand, in areas with lower employment rates, the challenge of space limitations is less obvious. However, access to appropriate resources and infrastructure is more difficult. It should be noted that, as shown in Figure 5, the employment rate is considered in three parameters namely Total Employment Rate, Male and Female Employment Rate. Males have the highest employment rate in the outskirts, such as Districts 17, 18, and 19. This is likely due to the majority of male-dominated industries like construction and manufacturing. However, the female employment rate is higher in the central Districts 6, 7, 11, and 12, possibly due to service-oriented activities and the higher number of female workers involved. By considering both male and female employment rates, along with population parameters, it becomes possible to identify and address gender inequalities in both employment access and the availability of green spaces. Considering socio-economic factors based on gender provides an accurate analysis of green space development. Urban planners and policymakers can move towards sustainable and fair development by addressing these problems.

Another parameter is the normalized parcel number. This parameter indicates urban density and provides insights into land use diversity and patterns of human activities. Due to the materials used in buildings, areas with higher building densities have larger impervious surfaces. The expansion of impervious surfaces, such as the development of asphalt and concrete materials, alters the energy balance and exacerbates the UHI effect [146]. On the other hand, the number and density of buildings are inversely correlated with access to green spaces [147]. In other words, areas with high densities face constraints in open spaces and green space development. For instance, in Figure 5, Districts 10, 11, 17, and 6 have high building densities and less green space (illustrated in Figure 6a). By considering the normalized parcel number, urban planners become aware of the open space limitation in different areas and should utilize modern technologies, such as green roofs and live walls, to enhance the quality of the urban ecosystem.

In addition, the dependency rate is an important socio-economic parameter. The dependency ratio measures how much a person relies on society for support and resources. In other words, they live off the society's goods but do not contribute any services or goods in return. The dependency ratio indicates the degree to which individuals rely on society for support. Simply, it reflects individuals who consume goods and services from society without actively contributing to its production or services in return. A region with more consumers and a smaller workforce typically exhibits a higher dependency rate. This rate can be calculated using two methods including net and gross. The gross dependency ratio is determined by dividing the population of inactive individuals (those aged under 15 and over 65) by the potential labor force (those aged 15-65). The net dependency ratio is the ratio of employed individuals to the total population. Figure 5h shows that Districts 16, 17, 18, and 19 have high dependency ratios, so the heavier burden on working people in these areas is obvious. In Districts 1, 2, 5, 6, 7, and 9, a more balanced net dependency rate, close to one, is observed. Conversely, Districts 3, 16, 17, 21, and 22 exhibit high gross dependency rates. These findings indicate a significant portion of the population in these districts is not actively engaged in the workforce. High dependency rates can significantly influence household consumption patterns. In areas with high dependency rates, the primary focus of household heads often shifts towards meeting basic life necessities. Consequently, investments in access to resources and green spaces and efforts to enhance overall quality of life are usually relegated to lower priorities.

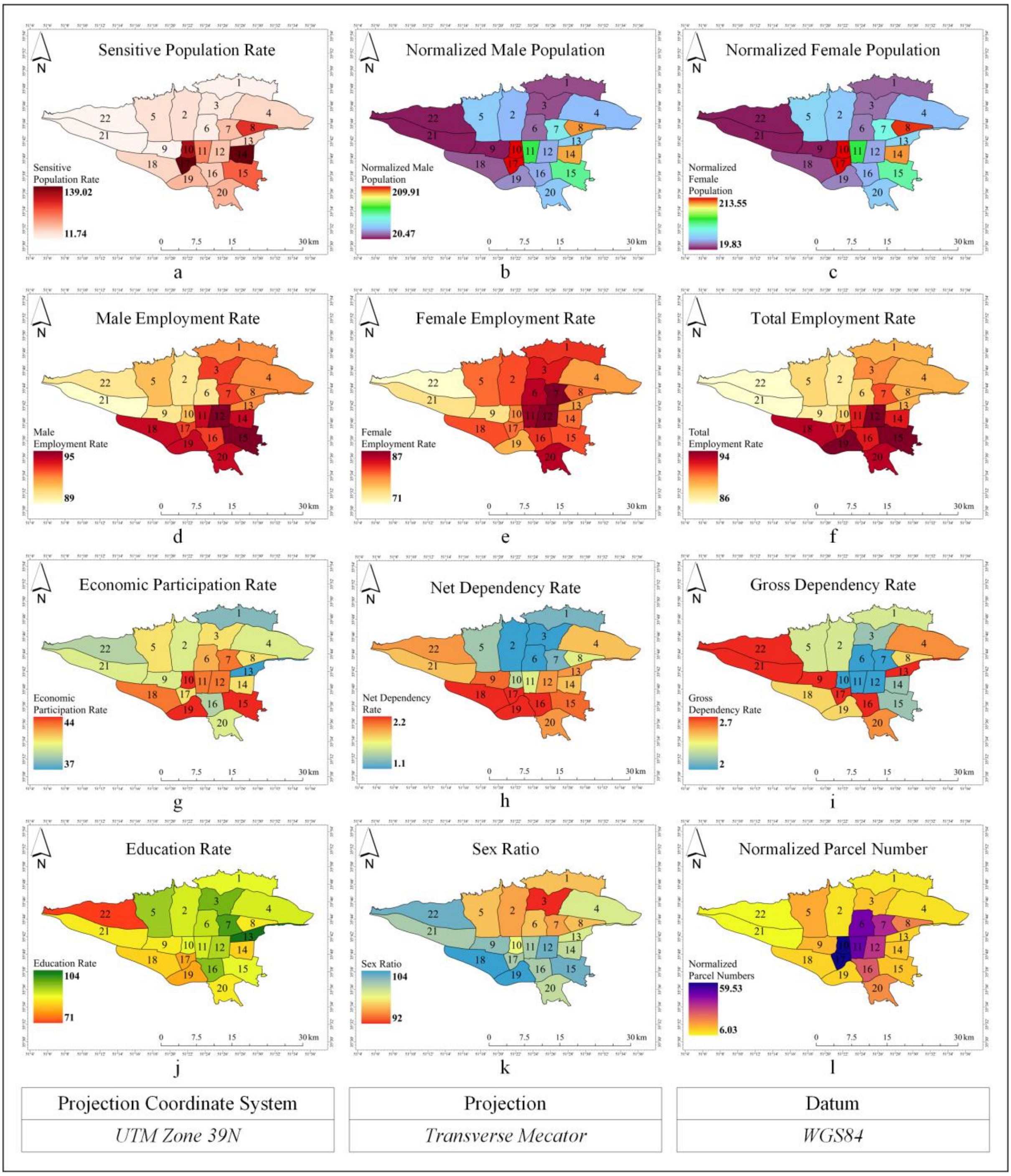

*Figure (5) The socio-economic alongside the sensitivity indices calculated for Tehran.*

## 5.2 Parameters Describing Urban Green Spaces

After performing the pre-processing on the raw data layers, Figure 6 displays the indices associated with green space in 22 districts of Tehran. One of these indices is the green belt areas, which refers to a continuous stretch of green space with a defined width, often developed as a forest at an appropriate distance from the urban boundary. As shown in Figure 6c, the green belt has developed more in the areas around Tehran. The next parameter is a green area, which includes the area of public green spaces. Another parameter is Green Corridors, which include tree planting, greenway trails, and forestry areas. In addition to the three aforementioned indices, the number and area of Tehran parks have been extracted and presented as indices LKC, LKM, L5KC, L5KM, L10KC, L10KM, L100KC, L100KM, M100KC, M100KM. Details of the aforementioned layers are provided in Table 1. These indices classify different levels of garden and park distribution in Tehran Municipality districts. For example, District 10 is considered to be the most densely populated area in Tehran, and its limited open space is one of the problems in the area. Similarly, there is a higher number of parks in this district than in others, however, the area of the parks is less than 1000 square meters (Figure 6). This information will help to make better decisions about the design and layout of urban green spaces.

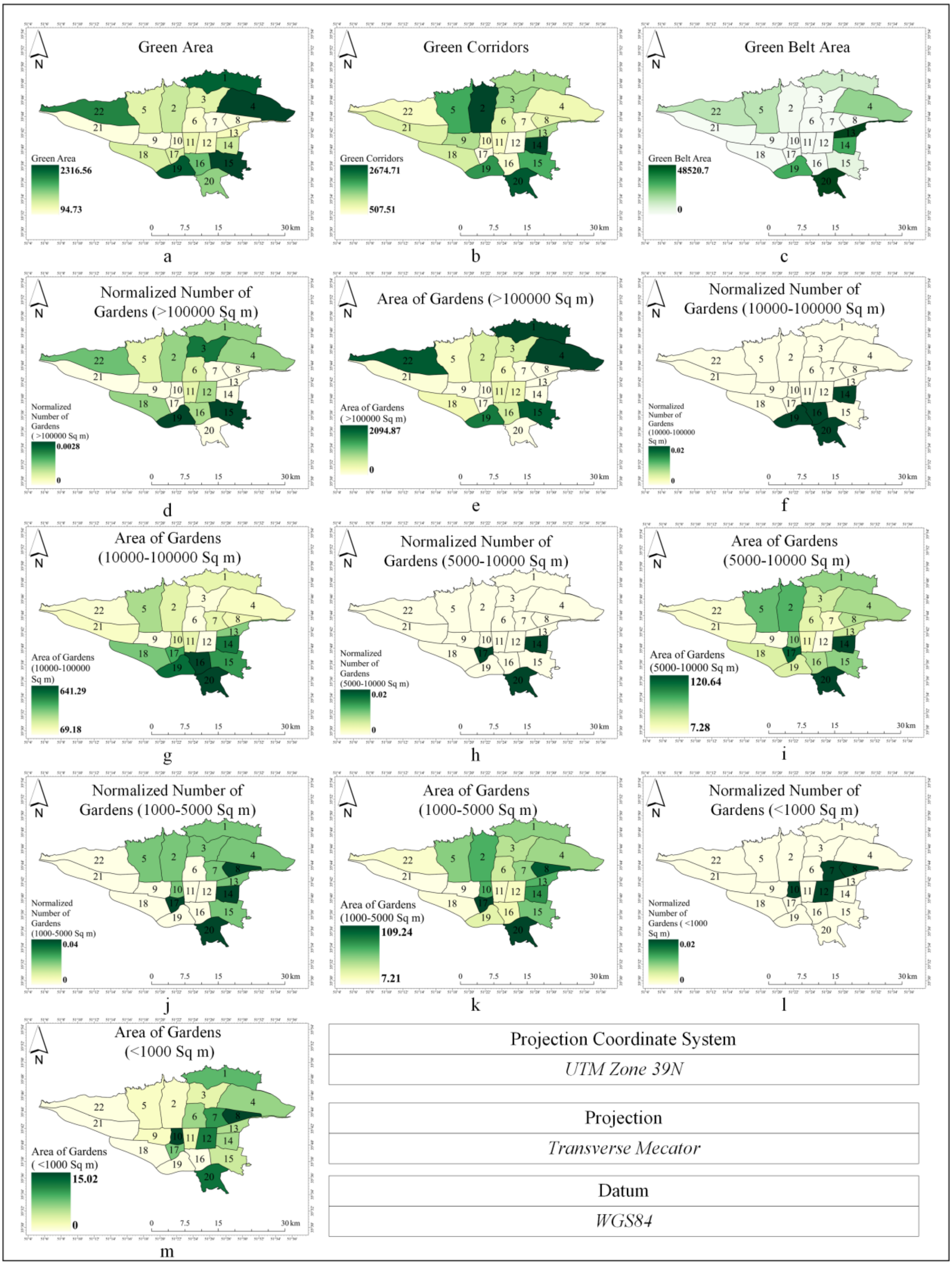

*Figure (6) Different parameters describing urban green space as well as the classification of parks based on their areas and numbers for Tehran.*

Integrating various green space parameters with diverse socio-economic indices contributes to a deeper understanding of the modeling process in subsequent steps. This integration aids in the development of a more comprehensive model for evaluating different areas and prioritizing green space development.

## 5.3 Result of $PM_{2.5}$ Concentration Modeling

The 10-year mean concentration map of $PM_{2.5}$ (2013-2023) was computed and can be observed in Figure 7a. The pollution level is higher in the southern and western parts of Tehran than in the northern and eastern regions, consistent with other research findings [148]. The minimum concentration level of $PM_{2.5}$ is 28 µg/m³, and the maximum is 35 µg/m³. Wind plays a crucial role in reducing $PM_{2.5}$ pollution. Lower wind speeds result in $PM_{2.5}$ pollutants remaining stationary, leading to increased concentration levels. In Tehran, the prevailing wind direction is from west to east. However, the construction of high-rise buildings can disrupt wind channels, hindering the dispersion of pollutants. On the other hand, land use changes and various construction projects in the west of Tehran have intensified the concentration of $PM_{2.5}$ in these areas. Also, according to Figure 5f, polluted areas like Districts 19 and 12 have higher employment rates. This issue highlights the link between various possible factors. It indicates that regions with higher employment rates are the center of industrial and commercial operations, where industrial processes, vehicle traffic, and commercial activities contribute to a significant increase in $PM_{2.5}$ levels. District 17 has high levels of $PM_{2.5}$. The Sensitive Population map in Figure 5a shows that this region has a highly sensitive population, which raises concerns regarding public health and environmental justice. To address this issue, effective interventions should be implemented, such as planning to identify polluting sources and protecting the health of citizens in this region. Referring to Figure 5, Districts 10, 11, and 17 exhibit high building densities, correlating with elevated concentrations of $PM_{2.5}$. These densely populated areas are often characterized by traffic congestion, which further exacerbates the accumulation of particles and pollutants. Areas with high building densities have limited green space, as indicated in the Green Area map in Figure 6a. This issue is important because vegetation cover can reduce $PM_{2.5}$ levels [149, 150], and consequently, the lack of green spaces leads to increased pollutant accumulation in these urban areas.

## 5.4 Result of WRF-based Simulation for T2 Estimation

According to Figure 3, three domains with resolutions of 9, 3, and 1 km have been defined in the WRF model simulation. It is necessary to validate the estimated T2 layer to ensure the model performance. The simulated temperature layer has been validated in resolutions of 3 km and 1 km (domains 2 and 3 in Figure 3) by 18 and 4 stations. The results of the validation can be seen in Table 3. Bias shows the model systematic errors. Negative bias at 3 km resolution indicates underestimation, while positive bias indicates overestimation at 1 km resolution. Mean Absolute Error (MAE) is an error metric that measures the average absolute difference between observations and estimations. The results show that the model is more accurate in the MAE metric at 1 km resolution. The Root Mean Square Error (RMSE) is another index that gives better results when the resolution is increased. In other words, the WRF model was able to generate T2 with an accuracy of less than 1°C using the nesting process at 1 km resolution. In general, Table 3 indicates that the model performs well in both nests. Moreover, fewer absolute errors have been obtained at a higher resolution.

*Table (3) The results of the statistical evaluations of T2 with the meteorological station observations.*

| Parameter | Resolution (Km) | RMSE (°C) | MAE (°C) | Bias (°C) |
|---|---|---|---|---|
| T2 | 3 | 1.75 | 1.39 | -1.06 |
| | 1 | 0.96 | 0.92 | 1.76 |

Figure 7b illustrates the map of the daily mean of T2 on the hottest day of the year (June 28, 2022). According to the WRF simulation results, T2 varies from 28°C to 36°C. Tehran southern and southwestern regions are hotter than the northern and northeastern areas, with more moderated temperatures. The T2 decrease in the northern and northeastern areas, such as Districts 1 and 4, is important for several reasons. Due to their higher altitude and proximity to cooler air masses, these areas experience lower temperatures caused by lower air pressure. As indicated in the Green Area map in Figure 6a and the spectral indices in Figure 8, Districts 1 and 4 exhibit a high percentage of vegetation. Green spaces and vegetation facilitate cooling by transferring moisture to the atmosphere through transpiration. In addition, vegetation can create shade and reduce the intensity of sunlight. The simulated temperature is forecasted to reach 36°C, posing a potential hazard for individuals vulnerable to heat stress. This could significantly impact the health of citizens, particularly sensitive groups. Referring to the sensitive population rate map in Figure 5a, Districts 10, 17, and 14, as well as Districts 11 and 15, exhibit a high percentage of sensitive populations. In addition, according to Figure 7b, these areas also experience higher temperatures. This issue indicates that it is necessary to develop green spaces and create easy access to medical centers, especially on hot days of the year in these areas. Also, in Figure 7b, the center of the city is warmer than the surroundings, representing the UHI effect. Based on the number of parcels map displayed in Figure 5, it can be seen that these central areas have a high population density. Based on vegetation indices derived from satellite imagery, Figure 8 also showed that these areas have lower vegetation coverage.

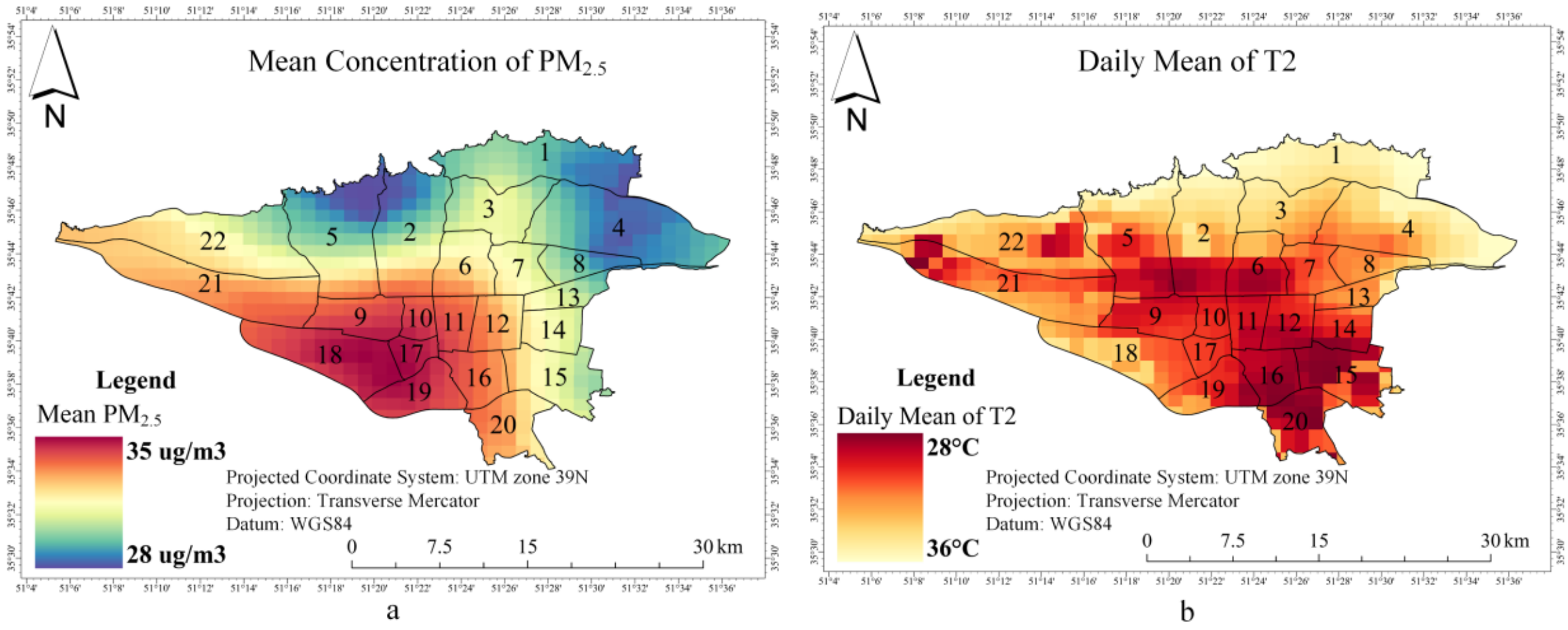

*Figure (7) The information layers produced for the Tehran include: a) the 10-year mean concentration of $PM_{2.5}$ (2013-21-03 to 2023-21-03), b) the daily mean of T2 simulated on the hottest day of 2022-06 -28 using the WRF model*

## 5.5 Result of LST and Spectral Indices Generation

The difference between daily and nightly LST is lower in central regions, according to Figure 8. The central parts of Tehran are denser and have more impervious surfaces. In this way, impervious surfaces absorb heat during the day and slowly release it at night. As a result, the night temperature becomes warmer, which leads to a decrease in the difference between daily and nightly LST. On the other hand, in the western districts, like Districts 22 and 21, there are a lower building density and more open spaces.

Since these regions have a weak absorption of heat, the temperature fluctuation between day and night can reach 22°C in some places.

Furthermore, a number of spectral indices were extracted to classify vegetation cover more precisely (Figure 8). Each of the indices is sensitive to various aspects of vegetation and exhibits different responses to different vegetation types. The utilization of multiple indices enables the extraction of a broader spectrum of plant characteristics, facilitating a more comprehensive evaluation in the classification process. In addition, integrating non-vegetation indices such as NDBI with NDMI can provide useful information for classifying vegetation from others. This becomes particularly important in areas with complex land use patterns.

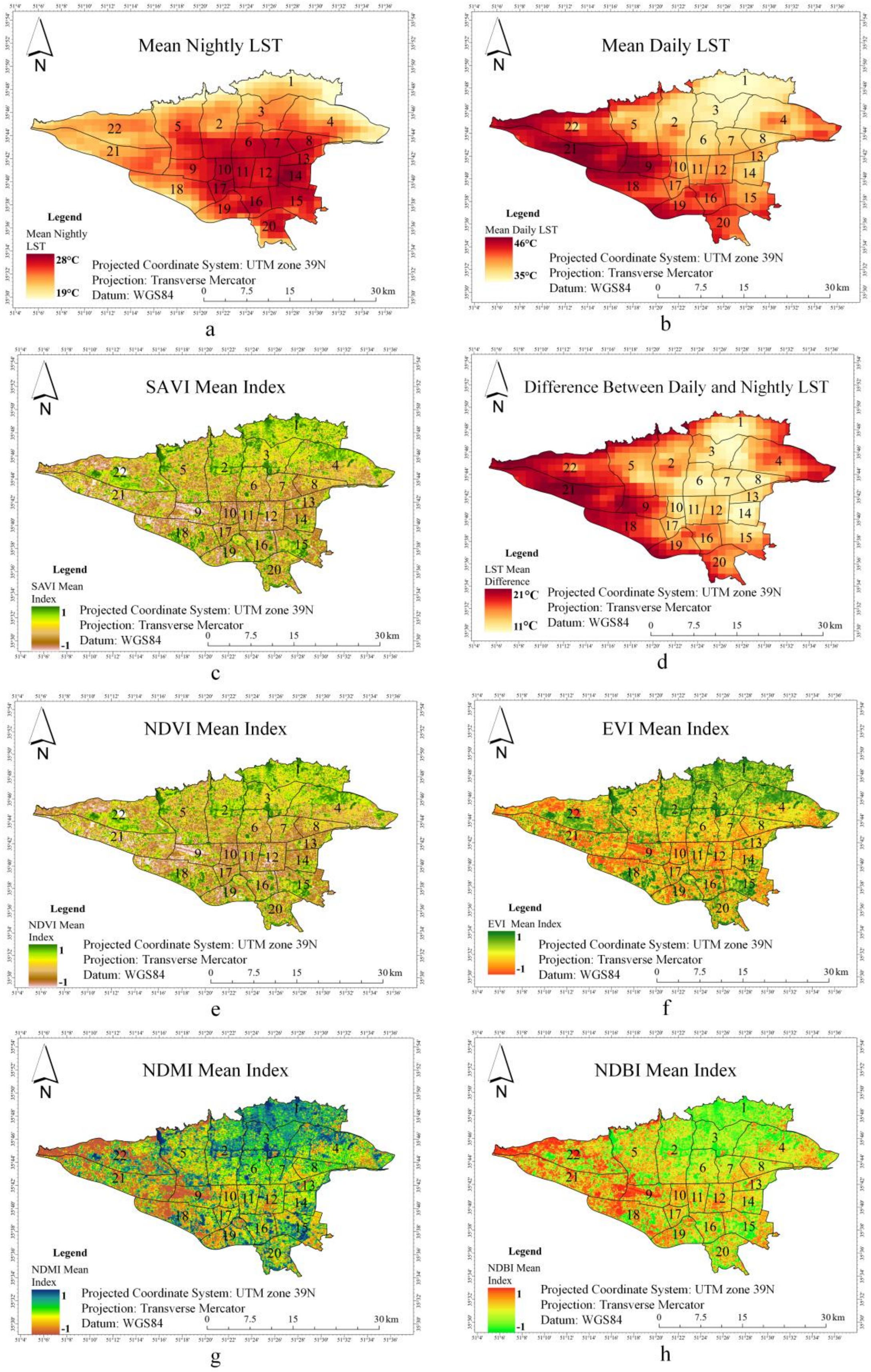

*Figure (8) Various spectral indices were calculated using GEE, including mean LST indices from MODIS on Terra satellite and mean spectral indices using Landsat 9 sensor.*

## 5.6 The Results of Correlation Analysis

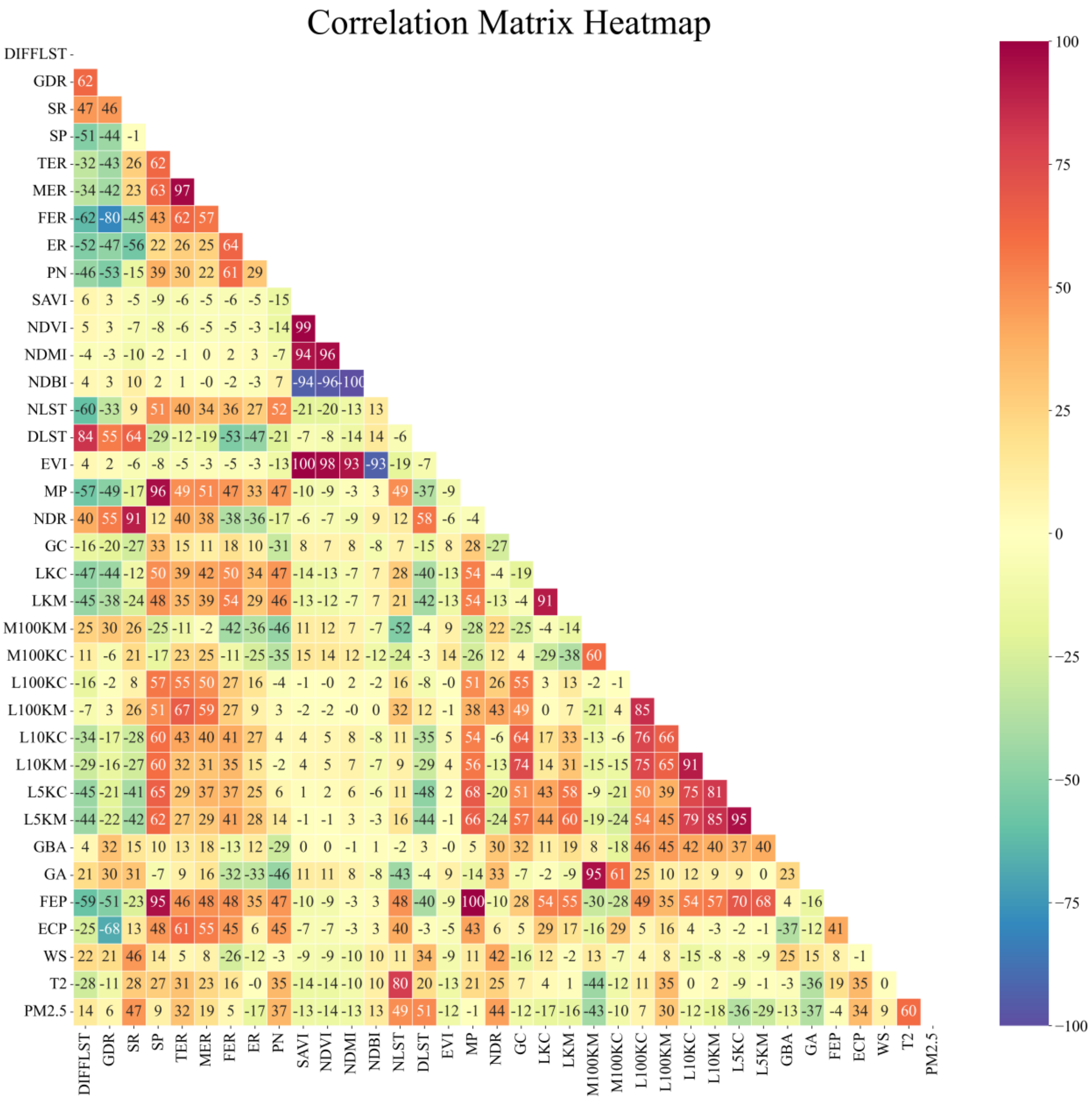

*Figure 9) A computed correlation matrix (in percentage) is used to evaluate the degree of similarity of different features*

By conducting correlation analysis, the degree of similarity among different features is determined. Figure 9 displays the result of this analysis. By this analysis, features with a similarity above 90% were excluded from the dataset. This threshold allows the retention of diverse indices, ensuring that the prioritization of areas is comprehensive and that important information is not lost. Out of the total 36 collected features, 12 features exhibited high similarity. Consequently, the final dataset consisted of 24 effective features. The 12 excluded features were SAVI, NDMI, EVI, NDBI, male and female population, net dependency rate, total employment rate, LKC, M100KM, L10KC, and L5KC indices.

In this process, only the NDVI index was retained among the various spectral indices. The other spectral indices were removed due to their high correlation with each other and the lack of effective additional information. Furthermore, male and female population indices were complementary to each other, with both being correlated with the sensitive population rate. However, the sensitive population rate and the sex ratio parameter were retained. Moreover, the net dependency rate is highly correlated with the sex ratio parameter. Consequently, this parameter was eliminated, given the presence of the gross dependency rate. Furthermore, the total employment rate was highly correlated with the male employment rate, leading to the removal of the former. It is worth noting that in this research, the employment rate was categorized into three types, including total, male, and female. After removing the total employment rate, the two other types remained for the analysis. Finally, among urban green space parameters, the number of LKC, L5KC, and L10KC parks correlated with their corresponding areas. Thus, the features related to the area were preserved. Furthermore, since the M100KM was highly correlated to the green area, subsequently, M100KM was removed from the dataset.

## 5.7 The Results of Binary Classification for Identifying the Vegetation Cover

To prioritize urban areas in green space development, it is essential to classify vegetated and non-vegetated areas. Subsequently, the non-vegetated areas should be prioritized. For classification purposes, after preparing the final dataset, hyperparameter optimization for the models was conducted. Following this, the performance of the models was evaluated using the identical test data. The results of this evaluation are presented in Table 4. Both the RF and XGBoost models exhibit the best performance. However, additional factors like ease of use and computational efficiency should also be considered. In general, all of the employed models have shown a consistent and high level of performance in identifying vegetated and non-vegetated areas. This suggests that the differences in criteria are minor, and all of the models are suitable for this purpose.

*Table (4) Evaluation of different classification models on test data using OA, Precision, Recall and F1-score criteria.*

| **Model** | **OA (%)** | **Precision (%)** | **Recall (%)** | **F1-score (%)** |
|---|---|---|---|---|
| **RF** | **94.00** | **94.03** | **94.00** | **94.00** |
| XGBoost | 94.00 | 94.00 | 94.00 | 94.00 |
| LightGBM | 93.58 | 93.62 | 93.58 | 93.58 |
| ET | 92.55 | 92.57 | 92.55 | 92.55 |

Compared to the XGBoost model, the RF model reveals better stability to outliers and noise, and the interpretation of feature importance is more straightforward. Therefore, the RF model was employed to generate a binary vegetation cover map (Figure 10). This map consists of two classes labeled with values 0 and 1. Class 0 denotes vegetated areas, whereas class 1 represents non-vegetated areas. Most of the vegetation in Tehran is located in the northern areas, while the central and southern regions have lower vegetation percentages due to human activities, limited open space, and air pollution.

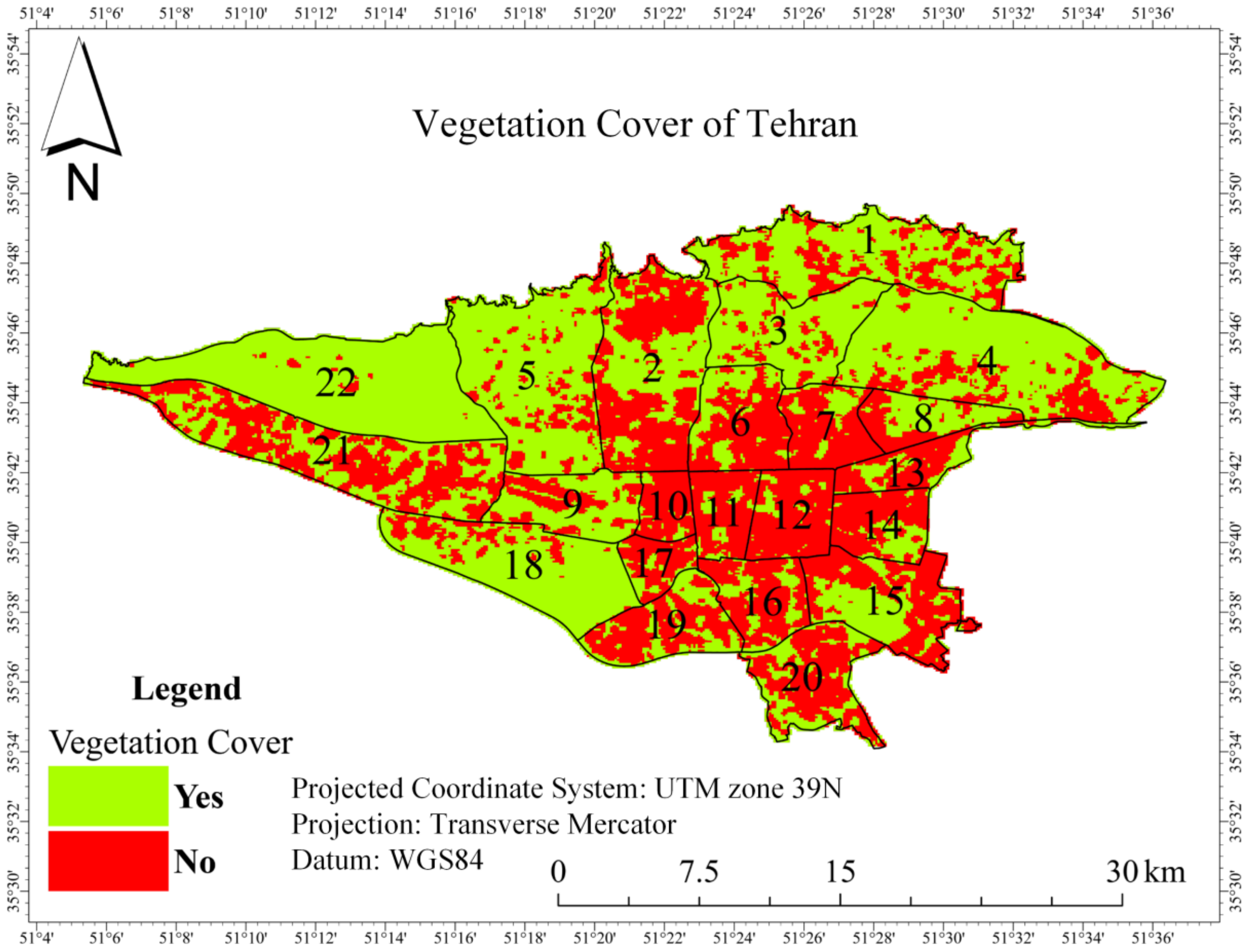

*Figure (10) A binary map of vegetation cover in Tehran indicating vegetated areas and non-vegetated areas.*

## 5.8 The Results of Priority Mapping of Green Space Development

As mentioned in Section 4.2, the NDVI, which primarily serves in binary vegetation cover classification, is removed from the dataset in order to emphasize the importance of environmental, socio-economic, and sensitivity indices in producing the priority map for green space development. Based on the new dataset, the RF model predicts the probability of being non-vegetated areas, consequently determining the degree of criticality for green space development. The findings presented in Table 5 indicate that the classification model, excluding the NDVI, has produced satisfactory results. However, there is still a discernible decrease in performance compared to the previous model (Table 4), including NDVI. As a result, the model was first used for binary classification, and then non-vegetated areas were prioritized.

*Table (5) The performance of RF model classification performance without NDVI index.*

| Model | OA (%) | Precision (%) | Recall (%) | F1-Score (%) |
|---|---|---|---|---|
| RF | 64.80 | 64.52 | 64.80 | 64.44 |

Figure 11 shows the priority map of green space development for Tehran. According to this map, Districts 1, 3, 4, 5, and 22, which have more green spaces and vegetation cover, include fewer critical areas, and if critical areas exist, their priority is low. On the other hand, the center of Tehran, such as Districts 7, 10, 11, 12, and 17, have more critical areas and require greater attention for green space development. Based on the normalized parcel number map in Figure 5, these districts have a high building density, leaving limited open space for urban greenery. Due to the high cost of land in central areas and limited government resources, creating urban green spaces by destroying buildings is impossible. Modern technologies related to vegetation cover, such as green roofs and living walls, are highly recommended in complex urban areas. These technologies have numerous benefits, including improving the environmental impact of buildings, reducing environmental issues such as the UHI effect and air pollution, and being economically sustainable [151, 152]. It is predicted that citizens of District 7, who have higher levels of education according to the education rate map in Figure 5, will exhibit better voluntary participation in sustainable development projects compared to other districts. Moreover, such projects improve people's quality of life, health, and thermal comfort.

It is possible to explore the critical areas based on the input indices. All these critical regions have had an unfavorable situation, at least based on one or more input indicators. In other words, a set of various environmental and socio-economic conditions and the sensitive population rate led to prioritizing non-vegetated areas based on their critical conditions. In District 10, a high-density area with many sensitive residents, citizens have experienced higher levels of $PM_{2.5}$. Also, this area has minimum green space, and its nightly LST is high.

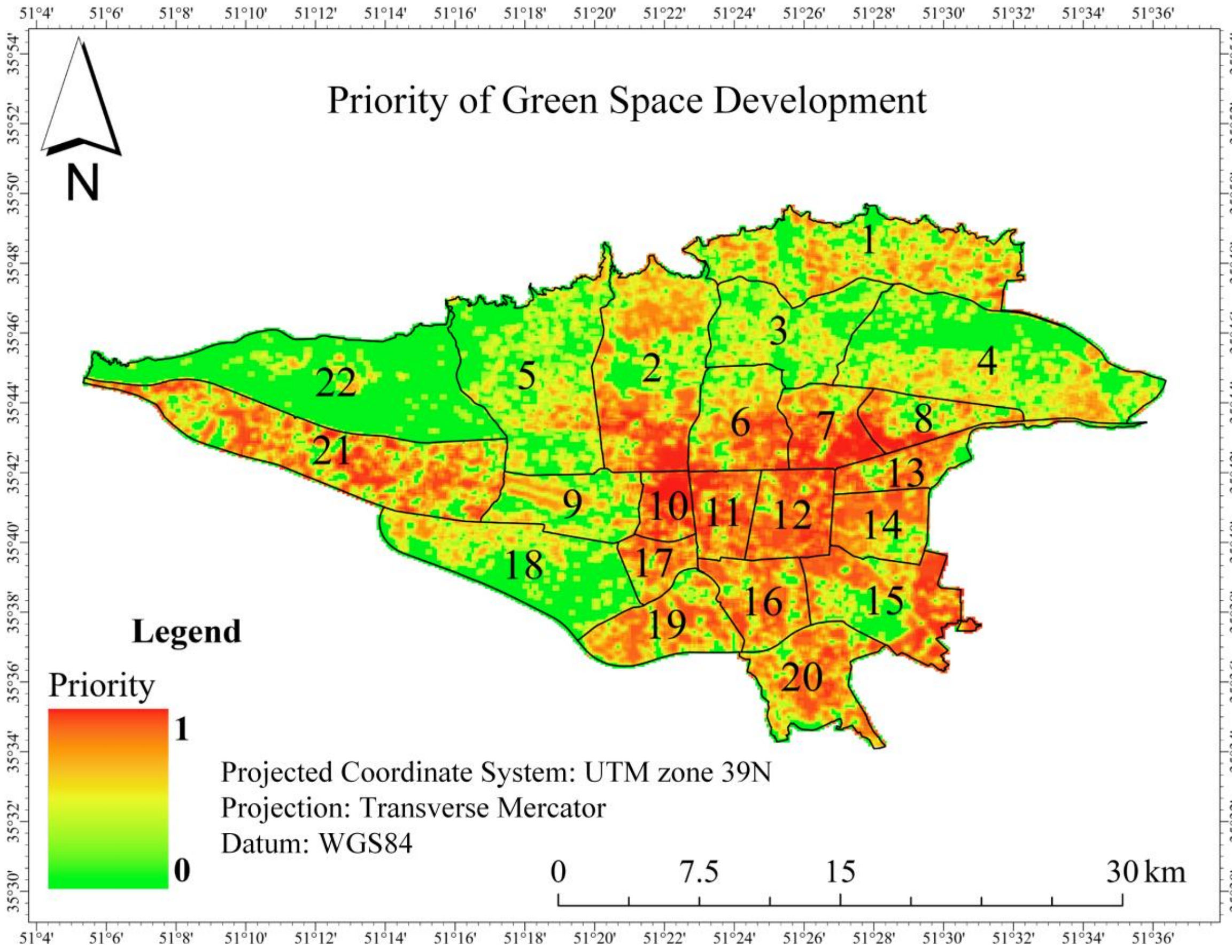

*Figure (11) The priority map of green space development by considering different socio-economic, environmental, and sensitivity indices.*

## 5.9 Feature Importance Analysis

The analysis of feature importance offers vital insights from the model. By analyzing feature importance, it becomes feasible to discern the contribution of various variables in the classification process. This enhances model interpretability and facilitates the identification of critical factors for prioritizing urban areas for green space development. Figure 12 shows the importance of the features in prioritizing the non-vegetated areas. Based on Figure 12, the most significant characteristic in determining the criticality of vegetation conditions is nightly LST (NLST). This is primarily due to its association with the UHI effect. Indeed, different urban areas exhibit varying thermal levels at night due to the types of impervious materials used in their construction. This distinction in nightly thermal levels is crucial in distinguishing between different areas. In other words, high nightly LST values indicate areas where impervious surfaces are typically dominant, and vegetation development would have a more significant cooling effect. Therefore, this parameter is important for urban green space planning, leading to improving urban comfort conditions. Following nightly LST, the sensitive population (SP) is ranked second. A highly sensitive population signifies areas where green space development is essential, as its improvement can provide numerous health benefits. The education rate (ER) is in the third place. This parameter can reflect people's participation level or citizens' awareness about their living place. Higher education levels represent more active communities in green space development. The next parameter is daily LST (DLST). Daily and nightly LST together offer a more comprehensive understanding of temperature patterns for the model. Considering that each area has a different thermal capacity, information on various thermal behavior will improve the classification accuracy. $PM_{2.5}$ and Wind Speed (WS) are ranked next with a minimum difference. Areas with high pollution levels tend to have poor vegetation, and $PM_{2.5}$ helps the model to understand these changes. Wind speed, on the other hand, affects temperature, vegetation, and buildings differently. Therefore, the type of vegetation and buildings in an area will have varying responses to wind speed. It is worth noting that the education rate, daily LST, $PM_{2.5}$, and wind speed indices exhibit very similar importance values. Each of these factors holds equal importance in the model performance. They all play vital roles in the classification process, indicating that the model behavior does not rely on a specific category of parameters. Instead, the model employes various socio-economic, environmental, and climatic indices to prioritize green space development.

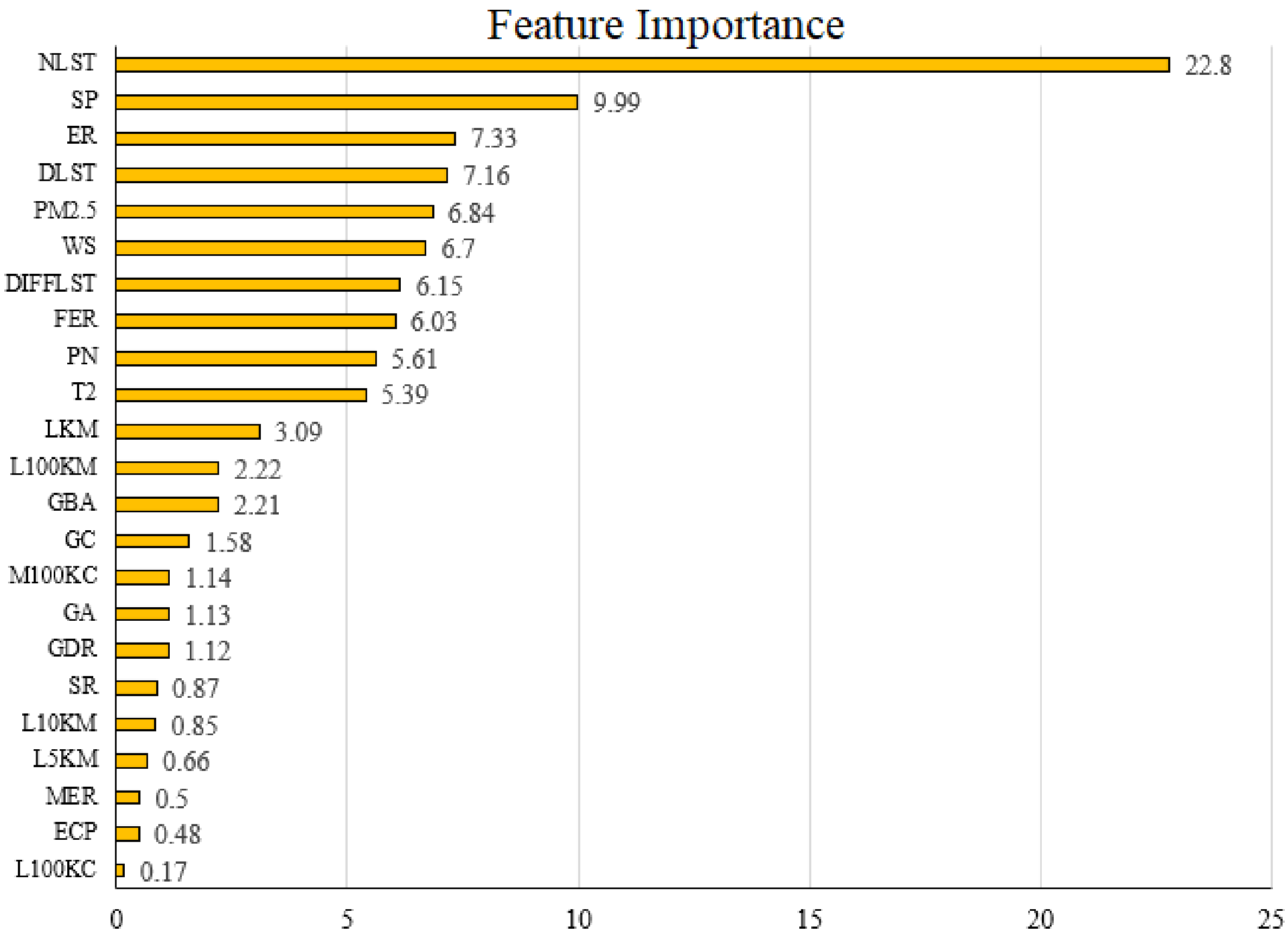

*Figure (12) The importance of features employed for generating the priority map of green space development.*

## 5.10 Microclimate-Based Validation of Urban Green Space Prioritization

To validate the achieved priority map, one of the most critical regions (200 m * 200 m) was selected to examine temperature conditions in the priority areas using microclimate simulations, based on the available data and processing power. Figure 13 depicts the selected critical area. Its temperature conditions were simulated as microclimate on the hottest day (2022-06-28), using Envi-Met software. In microclimate simulation, details such as building height and materials, soil type and plants, Digital Elevation Model (DEM), and land surface type are all essential. This pixel contains dense building blocks with varying heights. All necessary information was gathered in the field. In addition, in order to provide a suitable solution to improve temperature conditions, the scenario of using a green roof is evaluated. Green roof technology offers various environmental and economic benefits; however, it also faces challenges and drawbacks. Green roofs reduce energy consumption by 31-35% in the summer and by 2-10% in the winter, leading to cost savings [153, 154]. This technology also helps mitigate the UHI effect by absorbing solar radiation [155]. Furthermore, numerous studies have demonstrated the effectiveness of this technology in removing pollutants such as $CO_2$ and $NO_2$ [156, 157]. Economically, buildings with green roofs enhance aesthetic appeal and have higher property values [158]. They are also suitable in areas where traditional green space development, such as tree planting or park development, could not be more practical due to a lack of open space [159]. However, the construction of green roofs requires a significant initial investment, which can be a barrier for many property owners [160]. In addition, buildings utilizing this technology need to be reinforced to prevent additional loads from being imposed on them [161]. It is also worth noting that green roofs require periodic maintenance and repairs [162]. Considering all the advantages and disadvantages, and given that the identified critical area contains dense residential blocks where tree planting or other methods of vegetation development were not feasible, the impact of green roof technology has been evaluated as one of the common solutions for improving critical thermal conditions. Figure 14 illustrates two modeled scenarios.

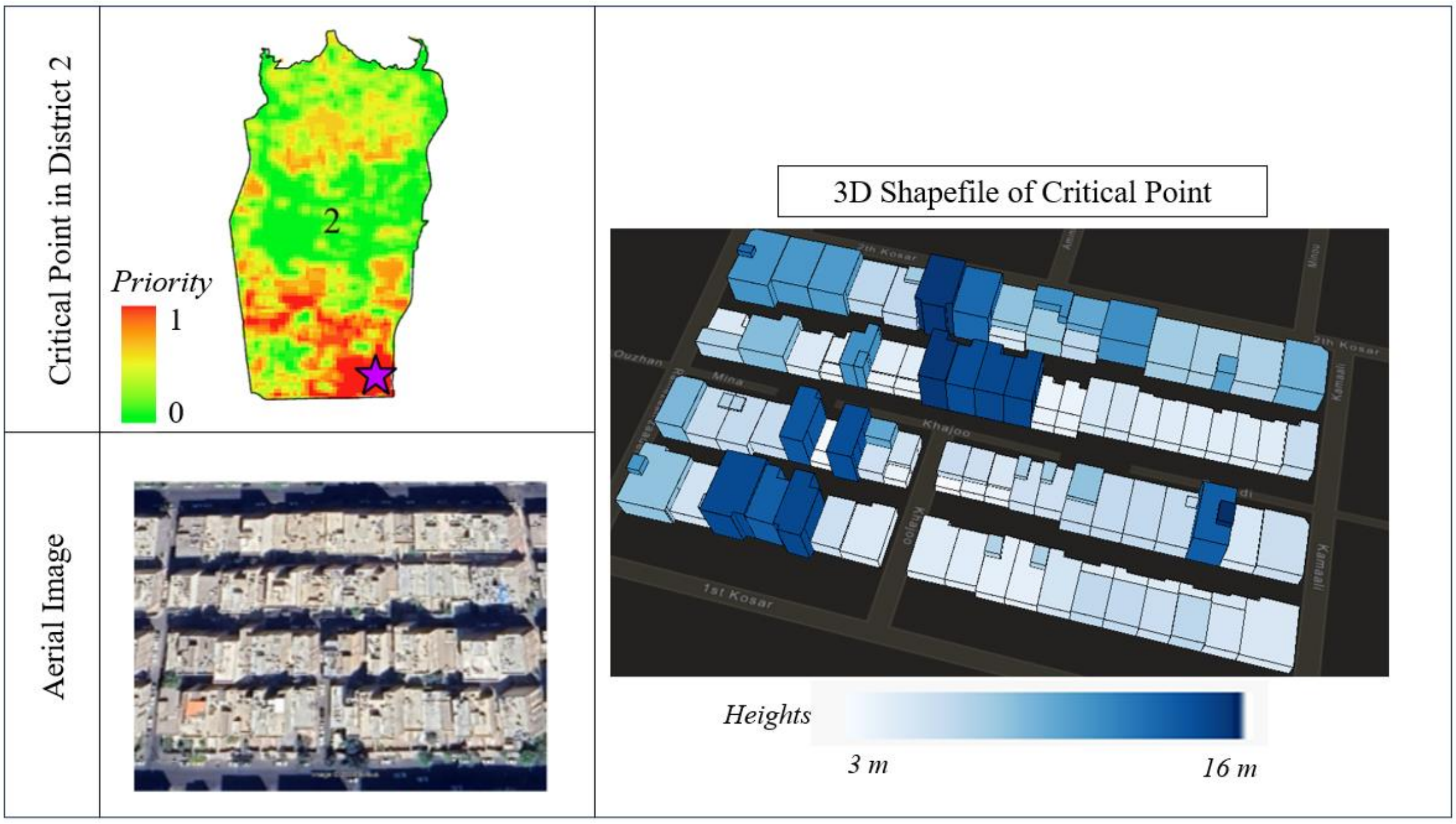

*Figure (13) A visualization of the selected critical area located in the District 2 of Tehran Municipality, by aerial images and the building height information layer.*

After completing the Envi-Met simulation, the air temperature was estimated for each scenario at a height of 1.80 m. By performing appropriate pre-processing, the amount of air temperature reduction caused by utilizing green roof technology was calculated at 15:00 which is visualized in Figure 14. The simulation results demonstrate that implementing green roof technology has reduced temperature conditions during peak heat hours. The maximum temperature reduction within the simulated range is 0.67°C, with an average reduction of 0.2°C overall. The Supplementary Materials provide detailed results from the microclimate simulations conducted for the two scenarios. Figure S1 presents maps illustrating temperature reductions resulting from the implementation of green roofs at 9:00 and 22:00. Furthermore, Figure S2 includes graphs depicting the maximum and average temperature reductions observed at different times throughout the day.

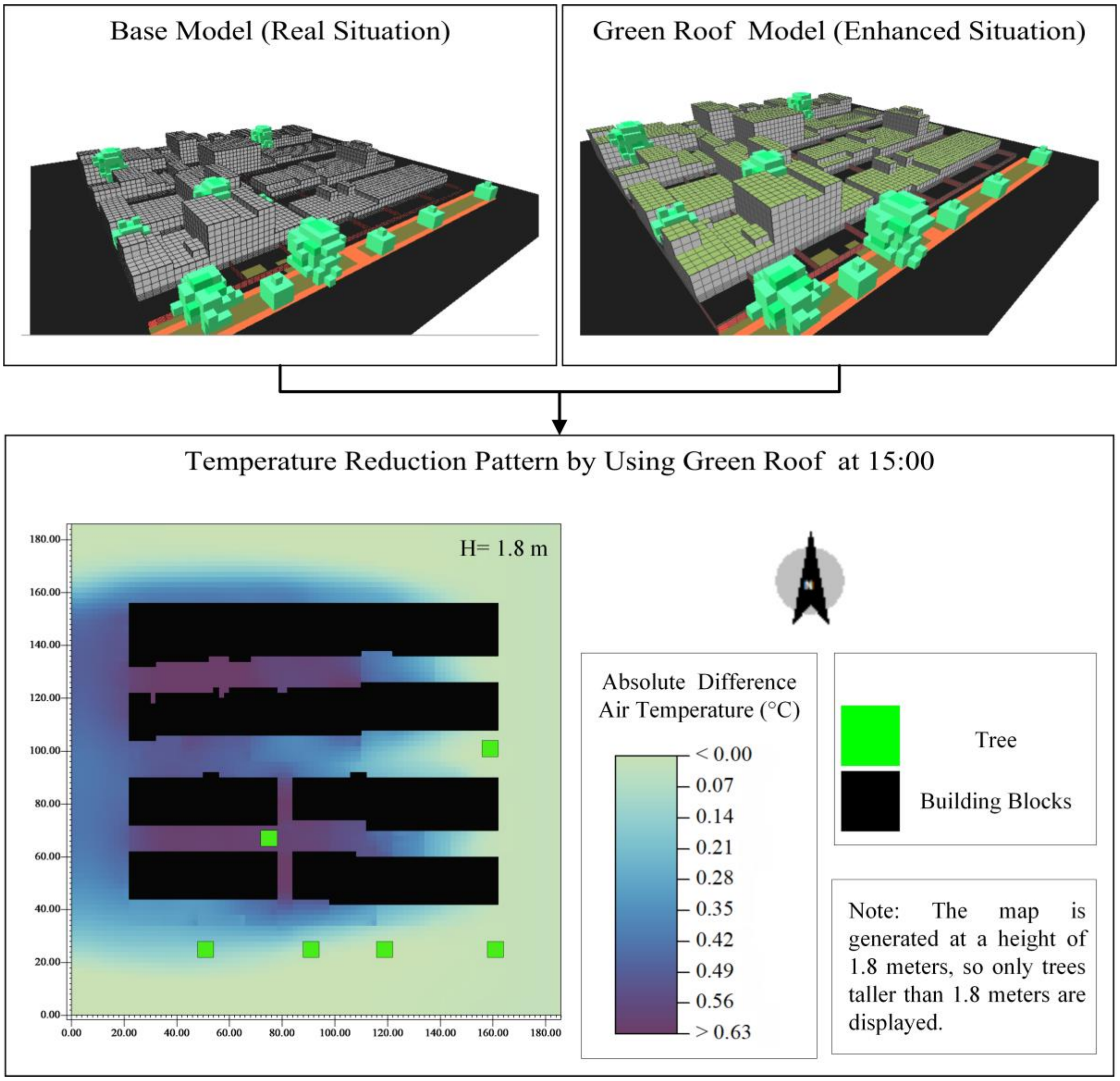

*Figure (14) Two scenarios for simulating microclimate conditions: 1) the real situation (Base Model) and 2) the implementation of a green roof (Green Roof Model). The results display the temperature reduction patterns at 15:00, measured at an altitude of 1.8 m, illustrating the cooling effect of the green roof on the urban environment.*

## 5.11 The Summary of the Results and Key Findings

Figure 15 summarizes the most important findings of this research. This study proposed a novel framework for generating a prioritization map for urban green space development by utilizing a wide range of socio-economic, environmental, and sensitivity indices based on ML algorithms. These indices were derived from resources such as statistical reports from Tehran Municipality, the GEE platform, meteorological stations, and the WRF modeling system, as illustrated in Figures 5, 6, 7, and 8. The performance of the WRF model was validated using data from meteorological stations, which demonstrated its ability to estimate air temperature at a resolution of 1 km with an accuracy of less than 1°C (RMSE of 0.96°C and MAE of 0.94°C). These findings suggest that numerical modeling can be an effective alternative in areas lacking sufficient meteorological stations. In this framework, a binary classification map of vegetation cover was generated using the RF model with an accuracy of 94.00%. This map was used first to identify areas lacking vegetation cover and subsequently prioritize them for green space development. Subsequently, by reapplying the feature selection process, a probability map indicating the likelihood of each area being critical due to the absence of vegetation cover was generated using the best model. The performance of the RF model in generating this map was evaluated using Feature Importance Analysis, as shown in Figure 12. The results indicated that nightly LST and the sensitive population index were the most significant indices among the others. In addition, the results of microclimate simulations conducted on one of the critical areas validated the framework performance. In a scenario, the impact of implementing green roofs on improving critical thermal conditions was assessed. The simulation results demonstrated the effectiveness of green roofs in reducing temperature by up to 0.67°C (Figure 14).

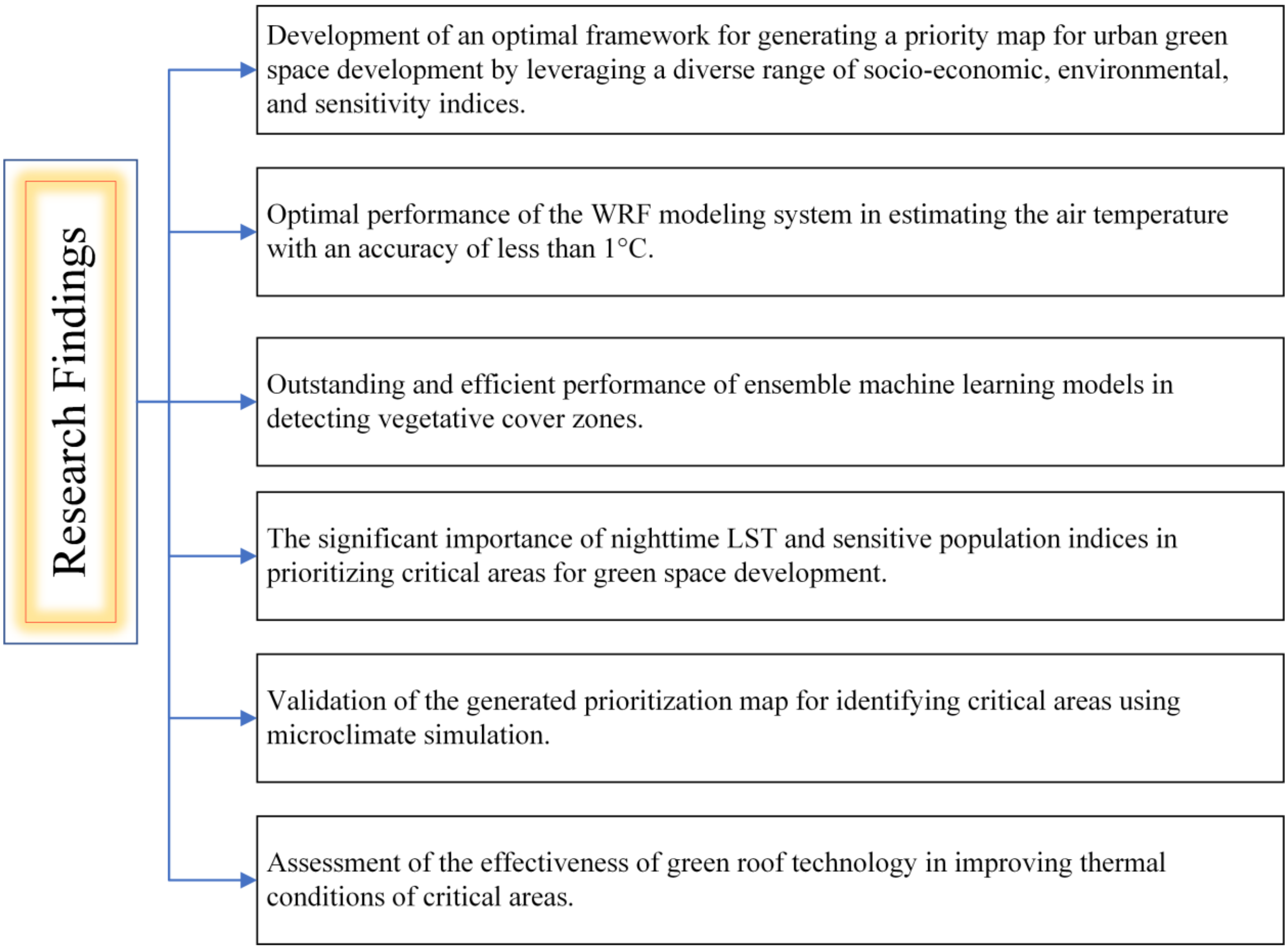

*Figure (15) Summary of the most important results obtained in this study.*

## 5.12 The Limitations and Future Directions

This study utilized the best available sources to prepare 36 socio-economic, environmental, and sensitivity indices. Since the indices were obtained from various sources, inconsistencies arose, particularly regarding spatial resolution. For example, despite the valuable information provided by the MODIS sensor regarding land surface temperature with a 1-day temporal resolution (both day and night), the spatial resolution of this sensor is not ideal. It is recommended that future studies explore the enhancement of the spatial resolution of LST products using statistical or dynamical downscaling. Consequently, it was essential to implement appropriate preprocessing steps to solve these kinds of inconsistencies among different indices. In this study, valid interpolation methods such as nearest neighbor were used to ensure all layers were resampled to a 100 m resolution and made them consistent before the classification process. Future studies may benefit from investigating more advanced downscaling and interpolation methods to achieve higher spatial resolutions to improve overall data quality for similar analyses.

While this research is aimed to utilize the most comprehensive features, it is suggested to involve additional indices such as building material, urban morphology parameters, immigrant rates, non-accidental death rates, and the number of hospital visitors on hot and pollutant days to enrich this framework. In other words, the framework has been developed based on the available datasets for Tehran. The overall process designed in this framework, such as integrating diverse socio-economic, environmental, and sensitivity indices, utilizing machine learning algorithms, and generating probability maps for critical vegetation conditions, can be adapted for other cities. However, the input data should be customized for each study to achieve better performance and reliability. By customizing the input data, it can be effectively adapted to different cities.

Also, it is important to note that some of the socio-economic and sensitivity indices were derived from statistical reports and data provided by Tehran Municipality. These indices, which reflect Tehran specific demographic, social, and economic characteristics, may not be equally relevant in other cities and might require adjustments or substitutions to align with local conditions. More precisely, the differences in the type and quality of available data, as well as the relative importance of the existing indices, are crucial factors in the application of this framework. For example, in Tehran, due to the lack of access to data about the distribution of migrants or the percentage of ethnic minorities, these indices were not directly incorporated into the analysis. Furthermore, Tehran cultural and social structure means that such indicators have limited relevance for prioritizing urban green space development. In contrast, in cities with high immigration levels or significant cultural diversity, indices like migrant density and the proportion of ethnic minorities could be critical in shaping green space development strategies. Therefore, considering local characteristics and available data is essential when applying this framework in other regions. These customizations of input layers may influence classification accuracy or increase uncertainty; however, the overall framework remains transferable. This framework can be applied in diverse geographical, social, and economic contexts by customizing the indices, making it a powerful tool for improving urban planning.

# 6 Conclusion

The development of green space in urban areas significantly impacts various aspects of life. However, identifying critical regions and prioritizing these regions is a highly challenging task for urban managers and planners. This issue is not a simple linear relationship, and various parameters are influential in determining the most critical areas for green space development. In previous research, knowledge-based methods have often been employed for weighting and prioritizing critical areas [52-55]. However, these methods heavily rely on expert opinions, which can introduce bias into the final map. Furthermore, precise prioritization in dynamic urban environments requires a broad range of indices, but previous studies often focused on limited environmental or socio-economic parameters. As the number of indices grows, involving experts across disciplines becomes challenging, increasing the risk of insufficient differentiation between indicators.

Another category of studies has used microclimate simulation to evaluate various scenarios for identifying the best locations for green space development [25, 26, 46-49]. Microclimate simulations require substantial computational resources and detailed data, often limiting their scope to small local areas. In addition, these studies typically prioritize vegetation development based solely on temperature or pollutant reduction, rarely incorporating socio-economic factors into the analysis. Thus, addressing the limitations of previous studies highlights the need for a comprehensive approach to urban green space prioritization. This approach should integrate a diverse range of indices and be applicable at a practical and suitable scale.

In this research, a novel framework was designed to prioritize urban green space development based on 36 different socio-economic, environmental, and sensitivity indices using ML algorithms. The application of ML allows the model to intelligently uncover hidden relationships within the data, minimizing potential bias. It is important to highlight that this framework leverages a range of data sources, including annual municipal reports, meteorological station data, air pollution monitoring, and the WRF model, to derive accurate and relevant indices.

The T2 layer, a key environmental component in this framework, was derived using WRF numerical modeling to address Tehran limited meteorological station coverage. Unlike interpolation-based statistical methods, WRF leverages numerical techniques and physical equations to estimate T2 with greater accuracy. Validation against meteorological measurements showed the model reliability, achieving RMSE and MAE values below 1°C on the hottest day of the year. Other environmental layers were extracted using the GEE, while social-economic and sensitivity indices were generated from municipal annual reports using ArcGIS Pro. Correlation analysis was applied to remove highly similar features (over 90%) from the dataset, reducing redundancy and preparing the final dataset for modeling. The first step in generating the urban green space prioritization map was vegetation cover classification. A total of 4,832 samples were collected via high-resolution Google Earth imagery and classified using machine learning models (RF, ET, LightGBM, XGBoost). RF outperformed others, achieving 94% accuracy, recall, and F1-score.

To prioritize areas without vegetation for green space development, NDVI was excluded from the dataset to emphasize socio-economic, environmental, and sensitivity indices. This approach guided the model to uncover hidden patterns, highlighting the criticality of various areas. More specifically, the RF model estimated the probability of being critical regarding vegetation cover for each area. Finally, by combining the binary map and the obtained probability map, a green space development prioritization map was created on a range from 0 to 1. In this map, a value of 0 indicates the presence of vegetation cover, signifying no priority for green space development, while a value of 1 represents areas with a high priority for development. This framework enabled the precise classification of vegetation cover. Subsequently, areas without vegetation were prioritized by incorporating a comprehensive set of socio-economic, environmental, and sensitivity indices. Then, the performance of the developed framework was further evaluated using Feature Importance Analysis and microclimate simulation. The Feature Importance Analysis validated the logical performance of the developed framework. The analysis showed that the nightly LST index is the most important, as it significantly influences UHI effects. The sensitive population rate ranked second, highlighting areas with vulnerable individuals, where green space development is a high priority for improving health and quality of life.

Furthermore, the importance of various features was balanced, indicating that the model generates the green space prioritization map using a combination of indices rather than a single one.

In a further experiment, the accuracy of the framework in identifying critical areas was evaluated using microclimate temperature simulations with Envi-Met. Furthermore, to provide a practical solution for mitigating critical temperature conditions in the identified areas, a scenario investigated the impact of green roof technology. The results showed that implementing green roofs can reduce temperatures by up to 0.67°C.

# 7 References

[1] T. W. Bank. "Urban Development." https://www.worldbank.org/en/topic/urbandevelopment/overview (accessed 2/29/2024, 2024).

[2] Y. Xiao, L. Guo, and W. Sang, "Impact of Fast Urbanization on Ecosystem Health in Mountainous Regions of Southwest China," *International Journal of Environmental Research and Public Health*, vol. 17, no. 3,

[3] N. Zhang, H. Ye, M. Wang, Z. Li, S. Li, and Y. Li, "Response Relationship between the Regional Thermal Environment and Urban Forms During Rapid Urbanization (2000–2010–2020): A Case Study of Three Urban Agglomerations in China," *Remote Sensing*, vol. 14, no. 15,

[4] J. L. Zambrano-Martinez, C. T. Calafate, D. Soler, J.-C. Cano, and P. Manzoni, "Modeling and Characterization of Traffic Flows in Urban Environments," *Sensors*, vol. 18, no. 7,

[5] J.-P. Kim and J.-M. Guldmann, "Land-Use Planning and the Urban Heat Island," *Environment and Planning B: Planning and Design,* vol. 41, no. 6, pp. 1077-1099, 2014.

[6] T. T. Van, N. D. H. Tran, H. D. X. Bao, D. T. T. Phuong, P. K. Hoa, and T. T. N. Han, "Optical Remote Sensing Method for Detecting Urban Green Space as Indicator Serving City Sustainable Development," in *Proceedings*, vol. 2, 3 ed.: MDPI, 2017, p. 140.

[7] R. G. Davies *et al.*, "City-Wide Relationships between Green Spaces, Urban Land Use and Topography," *Urban Ecosystems,* vol. 11, pp. 269-287, 2008.

[8] C. G. Threlfall *et al.*, "Increasing Biodiversity in Urban Green Spaces through Simple Vegetation Interventions," *Journal of Applied Ecology,* vol. 54, no. 6, pp. 1874-1883, 2017.

[9] F. Ferrini, A. Fini, J. Mori, and A. Gori, "Role of Vegetation as a Mitigating Factor in the Urban Context," *Sustainability,* vol. 12, no. 10, p. 4247, 2020.

[10] D. Gopal and H. Nagendra, "Vegetation in Bangalore's Slums: Boosting Livelihoods, Well-Being and Social Capital," *Sustainability,* vol. 6, no. 5, pp. 2459-2473, 2014.

[11] Shahfahad, M. Rihan, M. W. Naikoo, M. A. Ali, T. M. Usmani, and A. Rahman, "Urban Heat Island Dynamics in Response to Land-Use/Land-Cover Change in the Coastal City of Mumbai," *Journal of the Indian Society of Remote Sensing,* vol. 49, no. 9, pp. 2227-2247, 2021.

[12] Y. Hu, Z. Dai, and J.-M. Guldmann, "Modeling the Impact of 2d/3d Urban Indicators on the Urban Heat Island over Different Seasons: A Boosted Regression Tree Approach," *Journal of Environmental Management,* vol. 266, p. 110424, 2020.

[13] T. Basu and A. Das, "Urbanization Induced Degradation of Urban Green Space and Its Association to the Land Surface Temperature in a Medium-Class City in India," *Sustainable Cities and Society,* vol. 90, p. 104373, 2023/03/01/ 2023.

[14] H. Hou, Q. Longyang, H. Su, R. Zeng, T. Xu, and Z.-H. Wang, "Prioritizing Environmental Determinants of Urban Heat Islands: A Machine Learning Study for Major Cities in China," *International Journal of Applied Earth Observation and Geoinformation,* vol. 122, p. 103411, 2023/08/01/ 2023.

[15] M. C. Kondo, J. M. Fluehr, T. McKeon, and C. C. Branas, "Urban Green Space and Its Impact on Human Health," *International Journal of Environmental Research and Public Health,* vol. 15, no. 3, p. 445, 2018.
[16] J.-Y. Lee *et al.*, "Evidence-Based Field Research on Health Benefits of Urban Green Area," *Journal of the Korean Institute of landscape Architecture,* vol. 39, no. 5, pp. 111-118, 2011.
[17] R. C. Smardon, "Perception and Aesthetics of the Urban Environment: Review of the Role of Vegetation," *Landscape and Urban planning,* vol. 15, no. 1-2, pp. 85-106, 1988.
[18] R. Pandit, M. Polyakov, S. Tapsuwan, and T. Moran, "The Effect of Street Trees on Property Value in Perth, Western Australia," *Landscape and Urban Planning,* vol. 110, pp. 134-142, 2013.
[19] H. Sander, S. Polasky, and R. G. Haight, "The Value of Urban Tree Cover: A Hedonic Property Price Model in Ramsey and Dakota Counties, Minnesota, USA," *Ecological Economics,* vol. 69, no. 8, pp. 1646-1656, 2010.
[20] A. Gibson, R. Dodds, M. Joppe, and B. Jamieson, "Ecotourism in the City? Toronto's Green Tourism Association," *International Journal of Contemporary Hospitality Management,* vol. 15, no. 6, pp. 324-327, 2003.
[21] S. S. Clark and M. L. Miles, "Assessing the Integration of Environmental Justice and Sustainability in Practice: A Review of the Literature," *Sustainability*, vol. 13, no. 20,
[22] M. Gelobter, "The Meaning of Urban Environmental Justice," *Fordham Urb. LJ,* vol. 21, p. 841, 1993.
[23] V. Jennings, A. K. Baptiste, N. T. Osborne Jelks, and R. Skeete, "Urban Green Space and the Pursuit of Health Equity in Parts of the United States," *International Journal of Environmental Research and Public Health,* vol. 14, no. 11, p. 1432, 2017.
[24] R. Holifield, "Defining Environmental Justice and Environmental Racism," *Urban geography,* vol. 22, no. 1, pp. 78-90, 2001.
[25] A. MacLachlan, E. Biggs, G. Roberts, and B. Boruff, "Sustainable City Planning: A Data-Driven Approach for Mitigating Urban Heat," *Frontiers in Built Environment,* vol. 6, p. 519599, 2021.
[26] S. Teshnehdel, H. Akbari, E. Di Giuseppe, and R. D. Brown, "Effect of Tree Cover and Tree Species on Microclimate and Pedestrian Comfort in a Residential District in Iran," *Building and Environment,* vol. 178, p. 106899, 2020.
[27] H. Wang *et al.*, "The Effects of Tree Canopy Structure and Tree Coverage Ratios on Urban Air Temperature Based on Envi-Met," *Forests*, vol. 14, no. 1,
[28] C. Nyelele and C. N. Kroll, "A Multi-Objective Decision Support Framework to Prioritize Tree Planting Locations in Urban Areas," *Landscape and Urban Planning,* vol. 214, p. 104172, 2021.
[29] C. Li, T. Zhang, X. Wang, and Z. Lian, "Site Selection of Urban Parks Based on Fuzzy-Analytic Hierarchy Process (F-Ahp): A Case Study of Nanjing, China," *International Journal of Environmental Research and Public Health,* vol. 19, no. 20, p. 13159, 2022.
[30] Z. Saeedavi, B. Khalili Moghadam, M. Bagheri Bodaghabadi, and N. Rangzan, "Land Suitability Assessment for Urban Green Space Using Ahp and Gis: A Case Study of Ahvaz Parks, Iran," *Desert,* vol. 22, no. 1, pp. 117-133, 2017.
[31] K. Ziari and K. Zebardast, "Spatial Distribution and Equity of Urban Green Space Provision in Tehran Metropolis Using Hybrid Factor Analysis and Analytic Network Process (F′Anp) Model," *Geomatica,* vol. 76, no. 2, p. 100022, 2024/12/01/ 2024.
[32] N. Xu *et al.*, "Accurate Suitability Evaluation of Large-Scale Roof Greening Based on Rs and Gis Methods," *Sustainability*, vol. 12, no. 11,
[33] R. F. de Farias Aires and L. Ferreira, "A New Approach to Avoid Rank Reversal Cases in the Topsis Method," *Computers & Industrial Engineering,* vol. 132, pp. 84-97, 2019.
[34] M. L. Wilkerson *et al.*, "The Role of Socio-Economic Factors in Planning and Managing Urban Ecosystem Services," *Ecosystem Services,* vol. 31, pp. 102-110, 2018.
[35] L. Schebek and T. Lützkendorf, "Assessing Resource Efficiency of City Neighbourhoods: A Methodological Framework for Structuring and Practical Application of Indicators in Urban Planning," *Sustainability,* vol. 14, no. 13, p. 7951, 2022.

[36] G. Warth, A. Braun, O. Assmann, K. Fleckenstein, and V. Hochschild, "Prediction of Socio-Economic Indicators for Urban Planning Using Vhr Satellite Imagery and Spatial Analysis," *Remote Sensing,* vol. 12, no. 11, p. 1730, 2020.
[37] T. Banerjee, "Role of Indicators in Monitoring Growing Urban Regions the Case of Planning in India's National Capital Region," *Journal of the American Planning Association,* vol. 62, no. 2, pp. 222-235, 1996.
[38] B. K. L. Mak and C. Y. Jim, "Linking Park Users' Socio-Demographic Characteristics and Visit-Related Preferences to Improve Urban Parks," *Cities,* vol. 92, pp. 97-111, 2019.
[39] M. Jelokhani-Niaraki and J. Malczewski, "A Group Multicriteria Spatial Decision Support System for Parking Site Selection Problem: A Case Study," *Land Use Policy,* vol. 42, pp. 492-508, 2015.
[40] V. Sathyakumar, R. Ramsankaran, and R. Bardhan, "Linking Remotely Sensed Urban Green Space (Ugs) Distribution Patterns and Socio-Economic Status (Ses)-a Multi-Scale Probabilistic Analysis Based in Mumbai, India," *GIScience & Remote Sensing,* vol. 56, no. 5, pp. 645-669, 2019.
[41] F. De la Barrera, S. Reyes-Paecke, and E. Banzhaf, "Indicators for Green Spaces in Contrasting Urban Settings," *Ecological indicators,* vol. 62, pp. 212-219, 2016.
[42] D. Niemeijer and R. S. De Groot, "A Conceptual Framework for Selecting Environmental Indicator Sets," *Ecological indicators,* vol. 8, no. 1, pp. 14-25, 2008.
[43] M. S. Joy, P. Jha, P. K. Yadav, T. Bansal, P. Rawat, and S. Begam, "Site Suitability Analysis of Urban Green Parks in Ranchi City Using Gis–Ahp Based Multi-Criteria Decision Analysis," *Urbanization, Sustainability and Society,* vol. 1, no. 1, pp. 169-198, 2024.
[44] S. Guha, H. Govil, A. Dey, and N. Gill, "Analytical Study of Land Surface Temperature with Ndvi and Ndbi Using Landsat 8 Oli and Tirs Data in Florence and Naples City, Italy," *European Journal of Remote Sensing,* vol. 51, no. 1, pp. 667-678, 2018.
[45] B. Song and K. Park, "Contribution of Greening and High-Albedo Coatings to Improvements in the Thermal Environment in Complex Urban Areas," *Advances in Meteorology,* vol. 2015, no. 1, p. 792172, 2015.
[46] R. Li *et al.*, "Numerical Simulation Methods of Tree Effects on Microclimate: A Review," *Renewable and Sustainable Energy Reviews,* vol. 205, p. 114852, 2024.
[47] N. Meili *et al.*, "Tree Effects on Urban Microclimate: Diurnal, Seasonal, and Climatic Temperature Differences Explained by Separating Radiation, Evapotranspiration, and Roughness Effects," *Urban Forestry & Urban Greening,* vol. 58, p. 126970, 2021.
[48] D. Zhao *et al.*, "Role of Species and Planting Configuration on Transpiration and Microclimate for Urban Trees," *Forests,* vol. 11, no. 8, p. 825, 2020.
[49] N. J. Georgi and K. Zafiriadis, "The Impact of Park Trees on Microclimate in Urban Areas," *Urban Ecosystems,* vol. 9, pp. 195-209, 2006.
[50] N. Antoniou, H. Montazeri, M. Neophytou, and B. Blocken, "Cfd Simulation of Urban Microclimate: Validation Using High-Resolution Field Measurements," *Science of the Total Environment,* vol. 695, p. 133743, 2019.
[51] C. Li, X. Li, Y. Su, and Y. Zhu, "A New Zero-Equation Turbulence Model for Micro-Scale Climate Simulation," *Building and Environment,* vol. 47, pp. 243-255, 2012.
[52] E. Gelan, "Gis-Based Multi-Criteria Analysis for Sustainable Urban Green Spaces Planning in Emerging Towns of Ethiopia: The Case of Sululta Town," *Environmental Systems Research,* vol. 10, no. 1, p. 13, 2021/02/06 2021.
[53] D. Ozturk and F. Kılıç-Gul, "Gis-Based Multi-Criteria Decision Analysis for Parking Site Selection," *Kuwait Journal of Science,* vol. 47, no. 3, 2020.
[54] K. Aliniai, A. Yarahmadi, J. Z. Zarin, H. Yarahmadi, and S. B. Lak, "Parking Lot Site Selection: An Opening Gate Towards Sustainable Gis-Based Urban Traffic Management," *Journal of the Indian Society of Remote Sensing,* vol. 43, pp. 801-813, 2015.
[55] A. Zucca, A. M. Sharifi, and A. G. Fabbri, "Application of Spatial Multi-Criteria Analysis to Site Selection for a Local Park: A Case Study in the Bergamo Province, Italy," *Journal of Environmental Management,* vol. 88, no. 4, pp. 752-769, 2008/09/01/ 2008.

[56] A. Darko, A. P. C. Chan, E. E. Ameyaw, E. K. Owusu, E. Pärn, and D. J. Edwards, "Review of Application of Analytic Hierarchy Process (Ahp) in Construction," *International Journal of Construction Management,* vol. 19, no. 5, pp. 436-452, 2019.
[57] V. Podvezko, "Application of Ahp Technique," *Journal of Business Economics and management,* no. 2, pp. 181-189, 2009.
[58] L. Lamrini, M. C. Abounaima, F. Z. El Mazouri, M. Ouzarf, and M. T. Alaoui, "Mcdm Filter with Pareto Parallel Implementation in Shared Memory Environment," *Statistics, Optimization & Information Computing,* vol. 10, no. 1, pp. 192-203, 2022.
[59] E. Hontoria and N. Munier, "Uses and Limitations of the Ahp Method a Non-Mathematical and Rational Analysis," ed: Springer: Berlin/Heidelberg, Germany, 2021.
[60] F. Chajaei and H. Bagheri, "Machine Learning Framework for High-Resolution Air Temperature Downscaling Using Lidar-Derived Urban Morphological Features," *Urban Climate,* vol. 57, p. 102102, 2024/09/01/ 2024.
[61] H. Salehi and R. Burgueño, "Emerging Artificial Intelligence Methods in Structural Engineering," *Engineering structures,* vol. 171, pp. 170-189, 2018.
[62] R. Colombo, D. Bellingeri, D. Fasolini, and C. M. Marino, "Retrieval of Leaf Area Index in Different Vegetation Types Using High Resolution Satellite Data," *Remote Sensing of Environment,* vol. 86, no. 1, pp. 120-131, 2003.
[63] M. Gašparović, D. Medak, I. Pilaš, L. Jurjević, and I. Balenović, "Fusion of Sentinel-2 and Planetscope Imagery for Vegetation Detection and Monitoring," *The International Archives of the Photogrammetry, Remote Sensing and Spatial Information Sciences,* vol. 42, pp. 155-160, 2018.
[64] M. Ganjirad and H. Bagheri, "Google Earth Engine-Based Mapping of Land Use and Land Cover for Weather Forecast Models Using Landsat 8 Imagery," *Ecological Informatics,* vol. 80, p. 102498, 2024/05/01/ 2024.
[65] W. G. Su, F. Z. Su, and C. H. Zhou, "Virtual Satellite Construction and Application for Image Classification," in *IOP Conference Series: Earth and Environmental Science*, 2014 2014, vol. 17: IOP Publishing, 1 ed., p. 012084.
[66] G. B. Hall, N. W. Malcolm, and J. M. Piwowar, "Integration of Remote Sensing and Gis to Detect Pockets of Urban Poverty: The Case of Rosario, Argentina," *Transactions in GIS,* vol. 5, no. 3, pp. 235-253, 2001.
[67] D. N. M. Donoghue, "Remote Sensing: Environmental Change," *Progress in Physical Geography,* vol. 26, no. 1, pp. 144-151, 2002.
[68] S. K. Seelan, D. Baumgartner, G. M. Casady, V. Nangia, and G. A. Seielstad, "Empowering Farmers with Remote Sensing Knowledge: A Success Story from the Us Upper Midwest," *Geocarto International,* vol. 22, no. 2, pp. 141-157, 2007.
[69] M. Ganjirad and M. R. Delavar, "Flood Risk Mapping Using Random Forest and Support Vector Machine," *ISPRS Annals of the Photogrammetry, Remote Sensing and Spatial Information Sciences,* vol. 10, pp. 201-208, 2023.
[70] N. Gorelick, M. Hancher, M. Dixon, S. Ilyushchenko, D. Thau, and R. Moore, "Google Earth Engine: Planetary-Scale Geospatial Analysis for Everyone," *Remote Sensing of Environment,* vol. 202, pp. 18-27, 2017.
[71] I. Simonis, "Geospatial Big Data Processing in Hybrid Cloud Environments," in *Ieee International Geoscience and Remote Sensing Symposiu*: IEEE, 2018, pp. 419-421.
[72] J. Sun *et al.*, "An Efficient and Scalable Framework for Processing Remotely Sensed Big Data in Cloud Computing Environments," *IEEE Transactions on Geoscience and Remote Sensing,* vol. 57, no. 7, pp. 4294-4308, 2019.
[73] H. Tamiminia, B. Salehi, M. Mahdianpari, L. Quackenbush, S. Adeli, and B. Brisco, "Google Earth Engine for Geo-Big Data Applications: A Meta-Analysis and Systematic Review," *ISPRS Journal of Photogrammetry and Remote Sensing,* vol. 164, pp. 152-170, 2020.
[74] T. L. Griffiths, F. Callaway, M. B. Chang, E. Grant, P. M. Krueger, and F. Lieder, "Doing More with Less: Meta-Reasoning and Meta-Learning in Humans and Machines," *Current Opinion in Behavioral Sciences,* vol. 29, pp. 24-30, 2019.

[75] I. Roustaet al., "Spatiotemporal Analysis of Land Use/Land Cover and Its Effects on Surface Urban Heat Island Using Landsat Data: A Case Study of Metropolitan City Tehran (1988–2018)," *Sustainability*, vol. 10, no. 12,
[76] O. Alizadeh-Choobari, P. Ghafarian, and P. Adibi, "Inter-Annual Variations and Trends of the Urban Warming in Tehran," *Atmospheric Research,* vol. 170, pp. 176-185, 2016.
[77] H. Bagheri, "Using Deep Ensemble Forest for High-Resolution Mapping of Pm2. 5 from Modis Maiac Aod in Tehran, Iran," *Environmental Monitoring and Assessment,* vol. 195, no. 3, p. 377, 2023.
[78] K. Naddafi *et al.*, "Health Impact Assessment of Air Pollution in Megacity of Tehran, Iran," *Iranian Journal of Environmental Health Science & Engineering,* vol. 9, pp. 1-7, 2012.
[79] J. G. Masek *et al.*, "Landsat 9: Empowering Open Science and Applications through Continuity," *Remote Sensing of Environment,* vol. 248, p. 111968, 2020.
[80] H. You, X. Tang, W. Deng, H. Song, Y. Wang, and J. Chen, "A Study on the Difference of Lulc Classification Results Based on Landsat 8 and Landsat 9 Data," *Sustainability,* vol. 14, no. 21, p. 13730, 2022.
[81] X. Xiong, N. Che, and W. Barnes, "Terra Modis on-Orbit Spatial Characterization and Performance," *IEEE Transactions on Geoscience and Remote Sensing,* vol. 43, no. 2, pp. 355-365, 2005.
[82] S.-B. Duan *et al.*, "Validation of Collection 6 Modis Land Surface Temperature Product Using in Situ Measurements," *Remote Sensing of Environment,* vol. 225, pp. 16-29, 2019.
[83] L. Hu, N. A. Brunsell, A. J. Monaghan, M. Barlage, and O. V. Wilhelmi, "How Can We Use Modis Land Surface Temperature to Validate Long-Term Urban Model Simulations?," *Journal of Geophysical Research: Atmospheres,* vol. 119, no. 6, pp. 3185-3201, 2014.
[84] Z. Wan, Y. Zhang, Q. Zhang, and Z. L. Li, "Quality Assessment and Validation of the Modis Global Land Surface Temperature," *International Journal of Remote Sensing,* vol. 25, no. 1, pp. 261-274, 2004.
[85] C. R. Sunstein, "Behavioral Analysis of Law," *U. chI. l. rev.,* vol. 64, p. 1175, 1997.
[86] D. Schlosberg, "Reconceiving Environmental Justice: Global Movements and Political Theories," *Environmental politics,* vol. 13, no. 3, pp. 517-540, 2004.
[87] J. H. Kim, "Linking Land Use Planning and Regulation to Economic Development: A Literature Review," *Journal of Planning Literature,* vol. 26, no. 1, pp. 35-47, 2011.
[88] S. Gen, H. Shafer, and M. Nakagawa, "Perceptions of Environmental Justice: The Case of a Us Urban Wastewater System," *Sustainable Development,* vol. 20, no. 4, pp. 239-250, 2012.
[89] S. P. Semenov, V. V. Slavsky, M. V. Kurkina, A. O. Tashkin, O. V. Samarina, and A. A. Finogenov, "Computer Mathematical Models of Socio-Economic Systems Using Gis Technologies," *Yugra State University Bulletin,* vol. 17, no. 1, pp. 79-84, 2021.
[90] D. E. Pataki *et al.*, "Coupling Biogeochemical Cycles in Urban Environments: Ecosystem Services, Green Solutions, and Misconceptions," *Frontiers in Ecology and the Environment,* vol. 9, no. 1, pp. 27-36, 2011.
[91] J. R. Wolch, J. Byrne, and J. P. Newell, "Urban Green Space, Public Health, and Environmental Justice: The Challenge of Making Cities 'Just Green Enough'," *Landscape and Urban Planning,* vol. 125, pp. 234-244, 2014.
[92] N. Kabisch, S. Qureshi, and D. Haase, "Human–Environment Interactions in Urban Green Spaces—a Systematic Review of Contemporary Issues and Prospects for Future Research," *Environmental Impact assessment review,* vol. 50, pp. 25-34, 2015.
[93] M. Kermani, A. Jonidi Jafari, M. Gholami, A. Shahsavani, F. Taghizadeh, and H. Arfaeinia, "Ambient Air Pm2. 5-Bound Pahs in Low Traffic, High Traffic, and Industrial Areas Along Tehran, Iran," *Human and Ecological Risk Assessment: An International Journal,* vol. 27, no. 1, pp. 134-151, 2021.
[94] M. Yunesian, R. Rostami, A. Zarei, M. Fazlzadeh, and H. Janjani, "Exposure to High Levels of Pm2. 5 and Pm10 in the Metropolis of Tehran and the Associated Health Risks During 2016–2017," *Microchemical Journal,* vol. 150, p. 104174, 2019.
[95] S. Faridi *et al.*, "Long-Term Trends and Health Impact of Pm2. 5 and O3 in Tehran, Iran, 2006–2015," *Environment international,* vol. 114, pp. 37-49, 2018.

[96] H. Bagheri, "A Machine Learning-Based Framework for High Resolution Mapping of Pm2. 5 in Tehran, Iran, Using Maiac Aod Data," *Advances in Space Research,* vol. 69, no. 9, pp. 3333-3349, 2022.
[97] Y. M. Govaerts, M. M. Verstraete, B. Pinty, and N. Gobron, "Designing Optimal Spectral Indices: A Feasibility and Proof of Concept Study," *International Journal of Remote Sensing,* vol. 20, no. 9, pp. 1853-1873, 1999.
[98] Y. Y. Tang, Y. Lu, and H. Yuan, "Hyperspectral Image Classification Based on Three-Dimensional Scattering Wavelet Transform," *IEEE Transactions on Geoscience and Remote sensing,* vol. 53, no. 5, pp. 2467-2480, 2014.
[99] G. C. Sánchez, O. Dalmau, T. E. Alarcón, B. Sierra, and C. Hernández, "Selection and Fusion of Spectral Indices to Improve Water Body Discrimination," *IEEE Access,* vol. 6, pp. 72952-72961, 2018.
[100] V. Yaloveha, D. Hlavcheva, and A. Podorozhniak, "Spectral Indexes Evaluation for Satellite Images Classification Using Cnn," *Journal of Information and Organizational Sciences,* vol. 45, no. 2, pp. 435-449, 2021.
[101] A. K. Piyoosh and S. K. Ghosh, "Analysis of Land Use Land Cover Change Using a New and Existing Spectral Indices and Its Impact on Normalized Land Surface Temperature," *Geocarto International,* vol. 37, no. 8, pp. 2137-2159, 2022.
[102] R.-b. Xiao, Z.-y. Ouyang, H. Zheng, W.-f. Li, E. W. Schienke, and X.-k. Wang, "Spatial Pattern of Impervious Surfaces and Their Impacts on Land Surface Temperature in Beijing, China," *Journal of Environmental Sciences,* vol. 19, no. 2, pp. 250-256, 2007.
[103] Z. Wu and Y. Zhang, "Spatial Variation of Urban Thermal Environment and Its Relation to Green Space Patterns: Implication to Sustainable Landscape Planning," *Sustainability*, vol. 10, no. 7,
[104] P. Qi, Y. Cui, H. Zhang, S. Hu, L. Yao, and L. B. Li, "Evaluating Multivariable Statistical Methods for Downscaling Nighttime Land Surface Temperature in Urban Areas," *IEEE Access,* vol. 8, pp. 162085-162098, 2020.
[105] R. Bala, R. Prasad, and V. Pratap Yadav, "A Comparative Analysis of Day and Night Land Surface Temperature in Two Semi-Arid Cities Using Satellite Images Sampled in Different Seasons," *Advances in Space Research,* vol. 66, no. 2, pp. 412-425, 2020/07/15/ 2020.
[106] S. García, A. Fernández, and F. Herrera, "Enhancing the Effectiveness and Interpretability of Decision Tree and Rule Induction Classifiers with Evolutionary Training Set Selection over Imbalanced Problems," *Applied Soft Computing,* vol. 9, no. 4, pp. 1304-1314, 2009.
[107] A. Garde, A. Voss, P. Caminal, S. Benito, and B. F. Giraldo, "Svm-Based Feature Selection to Optimize Sensitivity–Specificity Balance Applied to Weaning," *Computers in Biology and Medicine,* vol. 43, no. 5, pp. 533-540, 2013.
[108] A. R. Ismail and A. A. Zarir, "Comparative Performance of Deep Learning and Machine Learning Algorithms on Imbalanced Handwritten Data," *International Journal of Advanced Computer Science and Applications,* vol. 9, no. 2, 2018.
[109] A. Mirzaei, H. Bagheri, and M. Sattari, "Data Level and Decision Level Fusion of Satellite Multi-Sensor Aod Retrievals for Improving Pm2. 5 Estimations, a Study on Tehran," *Earth Science Informatics,* vol. 16, no. 1, pp. 753-771, 2023.
[110] M. Halimi, M. Farajzadeh, and Z. Zarei, "Modeling Spatial Distribution of Tehran Air Pollutants Using Geostatistical Methods Incorporate Uncertainty Maps," *Pollution,* vol. 2, no. 4, pp. 375-386, 2016.
[111] A. J. Crawford, D. H. McLachlan, A. M. Hetherington, and K. A. Franklin, "High Temperature Exposure Increases Plant Cooling Capacity," *Current Biology,* vol. 22, no. 10, pp. R396-R397, 2012.
[112] J. M. Pereyda-González *et al.*, "High Temperature and Elevated Co2 Modify Phenology and Growth in Pepper Plants," *Agronomy,* vol. 12, no. 8, p. 1836, 2022.
[113] J. Ban *et al.*, "The Effect of High Temperature on Cause-Specific Mortality: A Multi-County Analysis in China," *Environment international,* vol. 106, pp. 19-26, 2017.
[114] Y. Onoda and N. P. R. Anten, "Challenges to Understand Plant Responses to Wind," *Plant Signaling & Behavior,* vol. 6, no. 7, pp. 1057-1059, 2011.

[115] S. Ravi and P. D'Odorico, "A Field-Scale Analysis of the Dependence of Wind Erosion Threshold Velocity on Air Humidity," *Geophysical Research Letters,* vol. 32, no. 21, 2005.
[116] P. L. Kinney, "Climate Change, Air Quality, and Human Health," *American journal of Preventive Medicine,* vol. 35, no. 5, pp. 459-467, 2008.
[117] R. A. Memon, D. Y. C. Leung, and C.-H. Liu, "Effects of Building Aspect Ratio and Wind Speed on Air Temperatures in Urban-Like Street Canyons," *Building and Environment,* vol. 45, no. 1, pp. 176-188, 2010.
[118] M. Mohan and A. P. Sati, "Wrf Model Performance Analysis for a Suite of Simulation Design," *Atmospheric research,* vol. 169, pp. 280-291, 2016.
[119] D. Carvalho, A. Rocha, M. Gómez-Gesteira, and C. Santos, "A Sensitivity Study of the Wrf Model in Wind Simulation for an Area of High Wind Energy," *Environmental Modelling & Software,* vol. 33, pp. 23-34, 2012.
[120] C. Davis *et al.*, "Prediction of Landfalling Hurricanes with the Advanced Hurricane Wrf Model," *Monthly Weather Review,* vol. 136, no. 6, pp. 1990-2005, 2008.
[121] S. Shimada and T. Ohsawa, "Accuracy and Characteristics of Offshore Wind Speeds Simulated by Wrf," *Sola,* vol. 7, pp. 21-24, 2011.
[122] L. Pan, Y. Liu, J. C. Knievel, L. Delle Monache, and G. Roux, "Evaluations of Wrf Sensitivities in Surface Simulations with an Ensemble Prediction System," *Atmosphere,* vol. 9, no. 3, p. 106, 2018.
[123] M. Sun *et al.*, "Evaluation of Flood Prediction Capability of the Wrf-Hydro Model Based on Multiple Forcing Scenarios," *Water,* vol. 12, no. 3, p. 874, 2020.
[124] Y. Hu *et al.*, "Optimization and Evaluation of So2 Emissions Based on Wrf-Chem and 3dvar Data Assimilation," *Remote Sensing,* vol. 14, no. 1, p. 220, 2022.
[125] K. Shamsaei *et al.*, "Coupled Fire-Atmosphere Simulation of the 2018 Camp Fire Using Wrf-Fire," *International Journal of Wildland Fire,* vol. 32, no. 2, pp. 195-221, 2023.
[126] *The Gis4wrf Plugin*. (2018).
[127] D. Meyer and M. Riechert, "Open Source Qgis Toolkit for the Advanced Research Wrf Modelling System," *Environmental Modelling & Software,* vol. 112, pp. 166-178, 2019.
[128] W. C. Skamarock *et al.*, "A Description of the Advanced Research Wrf Version 4," *NCAR tech. note ncar/tn-556+ str,* vol. 145, 2019.
[129] H. Shahbazi, R. Ganjiazad, V. Hosseini, and M. Hamedi, "Investigating the Influence of Traffic Emission Reduction Plans on Tehran Air Quality Using Wrf/Camx Modeling Tools," *Transportation Research Part D: Transport and Environment,* vol. 57, pp. 484-495, 2017/12/01/ 2017.
[130] S. Zeyaeyan, E. Fattahi, A. Ranjbar, M. Azadi, and M. Vazifedoust, "Evaluating the Effect of Physics Schemes in Wrf Simulations of Summer Rainfall in North West Iran," *Climate*, vol. 5, no. 3,
[131] S. Zhang, X. Li, M. Zong, X. Zhu, and D. Cheng, "Learning K for Knn Classification," *ACM Transactions on Intelligent Systems and Technology (TIST),* vol. 8, no. 3, pp. 1-19, 2017.
[132] W. Yang and W. Nam, "Data Synthesis Method Preserving Correlation of Features," *Pattern Recognition,* vol. 122, p. 108241, 2022.
[133] M. Belgiu and L. Drăguţ, "Random Forest in Remote Sensing: A Review of Applications and Future Directions," *ISPRS journal of photogrammetry and remote sensing,* vol. 114, pp. 24-31, 2016.
[134] T. Chen and C. Guestrin, "Xgboost: A Scalable Tree Boosting System," in *Proceedings of the 22nd acm sigkdd international conference on knowledge discovery and data mining*, 2016 2016, pp. 785-794.
[135] Z. Chen and W. Fan, "A Freeway Travel Time Prediction Method Based on an Xgboost Model," *Sustainability,* vol. 13, no. 15, p. 8577, 2021.
[136] P. Geurts, D. Ernst, and L. Wehenkel, "Extremely Randomized Trees," *Machine learning,* vol. 63, pp. 3-42, 2006.
[137] X. Shi, Y. Cheng, and D. Xue, "Classification Algorithm of Urban Point Cloud Data Based on Lightgbm," in *IOP Conference Series: Materials Science and Engineering*, 2019 2019, vol. 631: IOP Publishing, 5 ed., p. 052041.

[138] J. Yan *et al.*, "Lightgbm: Accelerated Genomically Designed Crop Breeding through Ensemble Learning," *Genome biology,* vol. 22, pp. 1-24, 2021.
[139] J. Wu, S. Chen, and X. Liu, "Efficient Hyperparameter Optimization through Model-Based Reinforcement Learning," *Neurocomputing,* vol. 409, pp. 381-393, 2020.
[140] M. Grothe, P. Nijkamp, and H. J. Scholten, "Monitoring Residential Quality for the Elderly Using a Geographical Information System," *International Planning Studies,* vol. 1, no. 2, pp. 199-215, 1996.
[141] M. Jelokhani-Niaraki, F. Hajiloo, and N. N. Samany, "A Web-Based Public Participation Gis for Assessing the Age-Friendliness of Cities: A Case Study in Tehran, Iran," *Cities,* vol. 95, p. 102471, 2019/12/01/ 2019.
[142] G. Godbey and M. Blazey, "Old People in Urban Parks: An Exploratory Investigation," *Journal of Leisure Research,* vol. 15, no. 3, pp. 229-244, 1983.
[143] C. Wen, C. Albert, and C. Von Haaren, "The Elderly in Green Spaces: Exploring Requirements and Preferences Concerning Nature-Based Recreation," *Sustainable Cities and Society,* vol. 38, pp. 582-593, 2018.
[144] R. S. Kovats, S. Hajat, and P. Wilkinson, "Contrasting Patterns of Mortality and Hospital Admissions During Hot Weather and Heat Waves in Greater London, Uk," *Occupational and Environmental Medicine,* vol. 61, no. 11, pp. 893-898, 2004.
[145] A. Reuben, G. W. Rutherford, J. James, and N. Razani, "Association of Neighborhood Parks with Child Health in the United States," *Preventive medicine,* vol. 141, p. 106265, 2020.
[146] S. Vujovic, B. Haddad, H. Karaky, N. Sebaibi, and M. Boutouil, "Urban Heat Island: Causes, Consequences, and Mitigation Measures with Emphasis on Reflective and Permeable Pavements," *CivilEng*, vol. 2, no. 2, pp. 459-484
[147] D. Alexander and R. Tomalty, "Smart Growth and Sustainable Development: Challenges, Solutions and Policy Directions," *Local Environment,* vol. 7, no. 4, pp. 397-409, 2002.
[148] O. Alizadeh-Choobari, A. A. Bidokhti, P. Ghafarian, and M. S. Najafi, "Temporal and Spatial Variations of Particulate Matter and Gaseous Pollutants in the Urban Area of Tehran," *Atmospheric Environment,* vol. 141, pp. 443-453, 2016/09/01/ 2016.
[149] J. Chen, L. Zhu, P. Fan, L. Tian, and R. Lafortezza, "Do Green Spaces Affect the Spatiotemporal Changes of Pm2. 5 in Nanjing?," *Ecological processes,* vol. 5, no. 1, pp. 1-13, 2016.
[150] C.-H. Lim, J. Ryu, Y. Choi, S. W. Jeon, and W.-K. Lee, "Understanding Global Pm2. 5 Concentrations and Their Drivers in Recent Decades (1998–2016)," *Environment international,* vol. 144, p. 106011, 2020.
[151] K. Perini and P. Rosasco, "Cost–Benefit Analysis for Green Façades and Living Wall Systems," *Building and Environment,* vol. 70, pp. 110-121, 2013.
[152] M. Radić, M. Brković Dodig, and T. Auer, "Green Facades and Living Walls—a Review Establishing the Classification of Construction Types and Mapping the Benefits," *Sustainability,* vol. 11, no. 17, p. 4579, 2019.
[153] Z. Azkorra-Larrinaga, N. Romero-Antón, K. Martin-Escudero, and G. Lopez-Ruiz, "Environmentally Sustainable Green Roof Design for Energy Demand Reduction," *Buildings,* vol. 13, no. 7, p. 1846, 2023.
[154] O. Saadatian *et al.*, "A Review of Energy Aspects of Green Roofs," *Renewable and sustainable energy reviews,* vol. 23, pp. 155-168, 2013.
[155] T. Susca, S. R. Gaffin, and G. R. Dell'Osso, "Positive Effects of Vegetation: Urban Heat Island and Green Roofs," *Environmental pollution,* vol. 159, no. 8-9, pp. 2119-2126, 2011.
[156] Y. Arbid, "Impact of Green Roof Plants on the Removal of Air Pollutants (No2/O3) and on the Photochemical Fate of Pesticides," Université Clermont Auvergne, 2021. [Online].
[157] J.-f. Li *et al.*, "Effect of Green Roof on Ambient Co2 Concentration," *Building and Environment,* vol. 45, no. 12, pp. 2644-2651, 2010.
[158] M. Manso, I. Teotónio, C. M. Silva, and C. O. Cruz, "Green Roof and Green Wall Benefits and Costs: A Review of the Quantitative Evidence," *Renewable and Sustainable Energy Reviews,* vol. 135, p. 110111, 2021.
[159] J. Park, Y. Shin, S. Kim, S.-W. Lee, and K. An, "Efficient Plant Types and Coverage Rates for Optimal Green Roof to Reduce Urban Heat Island Effect," *Sustainability,* vol. 14, no. 4, p. 2146, 2022.

[160] M. Blackhurst, C. Hendrickson, and H. S. Matthews, "Cost-Effectiveness of Green Roofs," *Journal of Architectural Engineering,* vol. 16, no. 4, pp. 136-143, 2010.

[161] S. Weiler and K. Scholz-Barth, *Green Roof Systems: A Guide to the Planning, Design, and Construction of Landscapes over Structure*. John Wiley & Sons, 2009.

[162] B. Pirouz, S. A. Palermo, and M. Turco, "Improving the Efficiency of Green Roofs Using Atmospheric Water Harvesting Systems (an Innovative Design)," *Water,* vol. 13, no. 4, p. 546, 2021.